\documentclass[10pt,twocolumn,letterpaper]{article}

\usepackage[pagenumbers]{iccv} 
\usepackage[accsupp]{axessibility}  
\usepackage{xcolor}
\usepackage{soul}

\usepackage[T1]{fontenc}

\definecolor{iccvblue}{rgb}{0.21,0.49,0.74}
\usepackage[pagebackref,breaklinks,colorlinks,allcolors=iccvblue]{hyperref}
\usepackage{multirow}
\usepackage{array} 

\title{CULTURE3D: A Large-Scale and Diverse Dataset of Cultural Landmarks and Terrains for Gaussian-Based Scene Rendering}

\author{
Xinyi Zheng$^{1*}$,
Steve Zhang$^{2*}$,
Weizhe Lin$^{2\dagger}$,
Aaron Zhang$^{1}$, \\
Walterio W. Mayol-Cuevas$^{1}$,
Yunze Liu$^{2\dagger}$,
Junxiao Shen$^{1,2\dagger}$
\\
{\tt\small \{wf24018, fan.zhang, walterio.mayol-cuevas, junxiao.shen\}@bristol.ac.uk},\\
{\tt\small \{steve.zhang, weizhe.lin, liuyzchina\}@memories.ai}\\
$^1$\textbf{University of Bristol}\quad
$^2$\textbf{Memories.ai Research}\\
$^{*}$Equal contribution.\\
$^{\dagger}$Corresponding Author
}

\begin{document}

\twocolumn[{
\maketitle
\begin{center}
  \includegraphics[width=\textwidth]{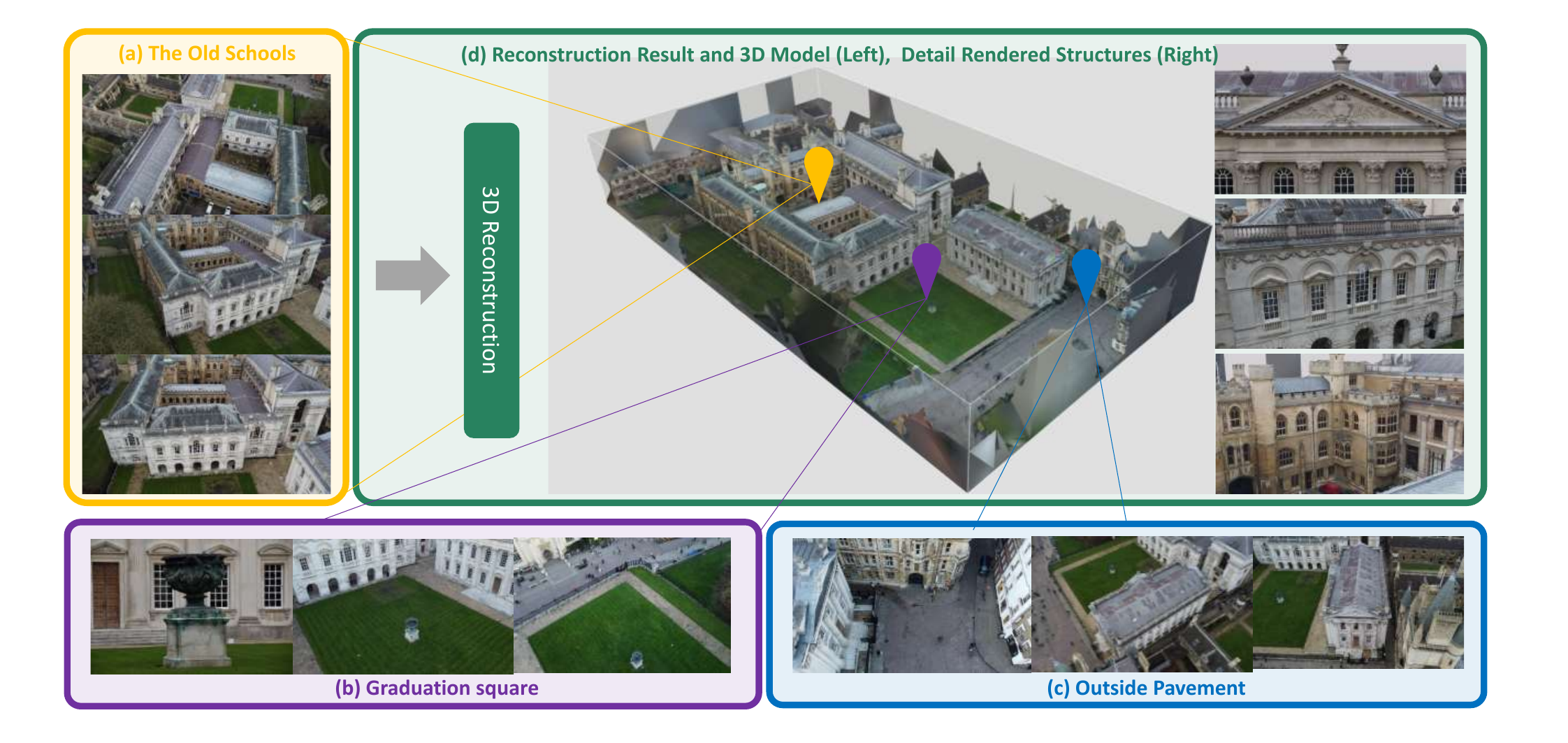}
  \vspace{-10pt}
  \captionof{figure}{Overview of the 3D reconstruction pipeline and resulting model. (a) Original images of The Old Schools building. (b) Original images of Graduation Square. (c) Original images of the outside pavement area. (d) Reconstructed 3D model integrating multiple spatial views, highlighting key locations within the reconstructed environment.}
  \label{fig:teaser}
\end{center}
}]

\begin{abstract}
Current state-of-the-art 3D reconstruction models face limitations in building extra-large scale outdoor scenes, primarily due to the lack of sufficiently large-scale and detailed datasets. In this paper, we present a extra-large fine-grained dataset with \textbf{10 billion} points composed of \textbf{41,006} drone-captured high-resolution aerial images, covering \textbf{20} diverse and culturally significant scenes from worldwide locations such as Cambridge Uni main buildings, the Pyramids, and the Forbidden City Palace. Compared to existing datasets, ours offers significantly larger scale and higher detail, uniquely suited for fine-grained 3D applications. Each scene contains an accurate spatial layout and comprehensive structural information, supporting detailed 3D reconstruction tasks. By reconstructing environments using these detailed images, our dataset supports multiple applications, including outputs in the widely adopted COLMAP format, establishing a novel benchmark for evaluating state-of-the-art large-scale Gaussian Splatting methods.
The dataset’s flexibility encourages innovations and supports model plug-ins, paving the way for future 3D breakthroughs. 
All datasets and code will be open-sourced for community use.

\end{abstract}    
\section{Introduction}
\label{sec:introduction}





The importance of large-scale 3D reconstruction has never been more pronounced, with applications spanning across numerous sectors including augmented reality, historical preservation, and urban planning. As we move towards more integrated digital and physical environments, the ability to accurately and efficiently map large areas in 3D is crucial. This technology not only enhances user experiences but also aids in the planning and management of complex infrastructure, making continued advancements in 3D reconstruction both essential and impactful.

Despite remarkable progress in the recent large-scale 3D scene reconstruction datasets, existing efforts like the GauU-Scene\cite{GauU-Scene}, MatrixCity-Aerial and MatrixCity-Street~\cite{matrixcity} datasets have their shortcomings. 
The synthetic nature of the MatrixCity datasets suffers from a significant domain shift from real-world scenarios, limiting their applicability in practical use cases that demand high-fidelity data.
Similarly, although GauU-Scene proves to be a useful resource, its scope is largely confined to relatively homogeneous scenes with an emphasis on outdoor environments, thereby limiting its applicability across a broader range of settings.
Moreover, existing datasets still exhibit limitations in both the quantity and diversity of scenes. In this paper, we introduce a larger-scale and higher-quality dataset to address these shortcomings.

Based on these observations, this paper introduces CULTURE3D, a dataset defined by its large-scale, high-resolution (48MP) imagery and diverse coverage of both indoor and outdoor environments. It not only meets the scale and quality for evaluating modern 3D reconstruction research but also offers greater diversity in scene styles, enhancing its applicability across various domains.
By integrating recordings from a wide range of geographical and architectural locations, this dataset aims to offer a robust benchmark that better reflects the complexity of real environments and meets the evolving needs of the technology and its applications. 

Specifically, CULTURE3D provides raw 2D image data and 3D models for 20 culturally significant scenarios, including historical landmarks (e.g., the Pyramids and Forbidden City), academic campuses (e.g., Cambridge University), religious sites, museums, and renowned architectural sites such as the Louvre Museum and Leaning Tower of Pisa.
We detail the process of data collection, the technologies used for 3D modeling, and the statistical properties of the dataset, such as the number of unique models, the range of environments covered, and the resolution of the data.

We further evaluated the effectiveness of our dataset by benchmarking various state-of-the-art Gaussian Splatting methods on CULTURE3D. Our findings, such as the notable differences in PSNR, SSIM, and LPIPS metrics across methods, reveal how different approaches perform under varied conditions presented by the new dataset. These results not only validate the quality and utility of CULTURE3D but also provide insights into the current capabilities and limitations of existing 3D reconstruction technologies, like out-of-memory error and failing under specific dataset scenes. These evaluations guide future research and development in large-scale cultural heritage scenes.

The contributions of this work are threefold as shown in the following:
\begin{itemize}
    \item \textbf{First Cultural Heritage High-Resolution Dataset}: CULTURE3D is the first publicly available large-scale dataset specifically built with ultra-high-resolution (48MP) drone imagery that spans diverse cultural and architectural landmarks worldwide, effectively bridging the gap between synthetic benchmarks and real-world complexity.
    \item \textbf{A Large-scale Scene Data Generation Pipeline:} We propose a comprehensive data collection and reconstruction pipeline designed to obtain high-quality, large-scale scene assets. Leveraging this pipeline, we present the most extensive and diverse large-scale scene dataset currently available, setting a new benchmark in the field.
    \item \textbf{Novel Benchmark for Advanced 3D Gaussian Splatting Methods}: We introduce detailed benchmarking using state-of-the-art 3D reconstruction methods, such as 3D Gaussian Splatting and Wild Gaussian, explicitly highlighting their limitations (e.g., out-of-memory errors and failure scenarios) when applied to large-scale, real-world cultural heritage environments.
\end{itemize}
\section{Related Work}
\label{sec:related_work}


\begin{table*}[t]
\centering
\scriptsize
\begin{tabular}{l l l l l l l}
\toprule
\textbf{Name} & \textbf{Year} & \textbf{Acquisition} & \textbf{Data Type} & \textbf{Area/Scale} & \textbf{Images} & \textbf{Points/Triangles} \\
\midrule
KITTI~\cite{KITTI}           & 2013 & Car Camera/LiDAR    & Image + LiDAR & -               & -     & 80K scans$^{\dagger}$ \\
TUM-RGBD~\cite{TUMVI}        & 2012 & Handheld RGB-D       & Image + Depth & Indoor labs      & -     & - \\
NCLT~\cite{NCLT}             & 2016 & Car Camera/LiDAR    & Image + LiDAR & Campus-scale     & -     & - \\
EuROC~\cite{EuROC}           & 2016 & Drone Camera         & Image         & Indoor rooms     & -     & - \\
DTU~\cite{DTU}               & 2016 & Static Camera        & Image         & Object-scale     & -     & Structured light \\
ScanNet~\cite{Dai2017ScanNetR3} & 2017 & Handheld RGB-D    & Image + Depth & Indoor scenes    & 2.5M  & 768K\,M$^{\dagger}$ \\
ETH3D~\cite{ETH3D}           & 2017 & Varied Cameras       & Image + LiDAR & Mixed scenes     & -     & FARO-based GT \\
Tanks \& Temples~\cite{Tanks_and_Temples} & 2017 & Static Camera & Image         & Mixed scenes     & -     & FARO-based GT \\
Complex Urban~\cite{ComplexUrban} & 2019 & UAV + GNSS-IMU  & PC + Image    & 0.7\,MP res.     & -     & - \\
WoodScape~\cite{woodscape}   & 2019 & Car Fisheye Cameras  & Image         & Street-scale     & -     & - \\
Newer College~\cite{newercollege} & 2020 & UAV + LiDAR     & PC + Image    & Outdoor campus   & -     & - \\
Hilti SLAM Challenge~\cite{Hilti_SLAM_Challenge} & 2022 & Varied & PC + Image & Mixed Scenes & - & - \\
LaMAR~\cite{LaMAR}           & 2022 & LiDAR + SfM          & PC + Image    & Mixed Scenes     & -     & - \\
ScanNet++~\cite{scannetplus} & 2023 & Handheld RGB-D       & Image + Depth & Indoor scenes    & 2.8\,MP res. & - \\
Hilti NSS~\cite{Nothing_Stands_Still} & 2023 & Matterport RGB-D & PC + Image & Indoor scenes & - & - \\
Oxford Spires~\cite{OxSpires}& 2024 & Leica RTC360         & PC + Image    & Mixed Scenes     & -     & - \\

{ModelNet10}~\cite{ModelNet40} & 2014 & Synthetic CAD & Mesh         & Object-level     & 4,899 & 10 classes \\
{ShapeNet}~\cite{ShapeNet}   & 2015 & Synthetic CAD & Mesh         & Object-level     & 51,190& 55 classes \\
{S3DIS}~\cite{s3dis}         & 2016 & Terrestrial LiDAR & PC       & Indoor building  & -     & 273M points \\
{Semantic3D}~\cite{Semantic3D} & 2017 & Terrestrial LiDAR & PC     & Outdoor sites    & -     & 4,009M points \\
{SemanticKITTI}~\cite{Behley2019SemanticKITTIAD} & 2019 & Car LiDAR & PC & City-scale & - & 4,549M points \\
{MatrixCity}~\cite{matrixcity} & 2023 & UAV/Vehicle LiDAR & PC + Image & City/Street-scale & 519k & - \\
{GauU Scene}~\cite{gauu_scene} & 2024 & UAV Ptgy       & PC + Image    & 6.6K\,MP$^{\dagger}$ & 46,000 & 628M points\\
\midrule
\textbf{Culture3D (Ours)} & 2025 & UAV Ptgy + RGB-D & PC + Image + Depth & Mixed Scenes & 41,006 & 10B points \\
\bottomrule
\end{tabular}

\vspace{2mm}
{\raggedright
\scriptsize
\caption{Merged overview of 3D scene datasets (including reconstruction and point cloud benchmarks). ``Ptgy'' stands for photogrammetry, a non-LiDAR data acquisition method. Only real world datasets are listed here, and some values are approximate.
\textbf{Notes:} 
- ``PC'' stands for point cloud. 
- ``Ptgy'' indicates photogrammetry (non-LiDAR). 
- $^{\dagger}$Approximate or reported figure. }
\label{tab:merged_overview}
\par}
\end{table*}

\subsection{Datasets for Small-scale 3D Reconstruction}

3D scene datasets provided benchmarks for environment understanding and semantic perception in indoor environments. These corpora facilitated foundational scene understanding across diverse tasks and established robust evaluation protocols. Stanford’s S3DIS \cite{armeni20163d} introduced semantic scans of six office areas; Matterport3D \cite{chang2017matterport3d} added 90 building-scale scenes with 194 k RGB-D images and object annotations for navigation and segmentation tasks; and ScanNet \cite{dai2017scannet} supplied over 2.5 M views, camera poses, and instance-level segmentations. However, these datasets focus mainly on homes and offices. As reconstruction research advanced, these indoor benchmarks guided the development and assessment of new algorithms. RGB-D SLAM collections such as Matterport3D \cite{Chang2017Matterport3DLF} and ScanNet \cite{Dai2017ScanNetR3} offered dense meshes, while Replica \cite{straub2019replica} improved reconstruction via simulation and rendering. Benchmark suites like TUM RGB-D (2012) \cite{Sturm2012EvaluatingEA} and sensor-fusion datasets EuROC \cite{EuROC} and TUM VI \cite{TUMVI} integrated visual–inertial data for enhanced localization and reconstruction. Unlike these mostly indoor, small-scale corpora, our Culture3D covers diverse large-scale indoor and outdoor scenes with higher fidelity and broader applicability.

\subsection{Datasets for Large-scale 3D Reconstruction}

Early 3D scene datasets were primarily developed for environmental understanding, laying the foundation for semantic 3D perception. In outdoor environments, initial reconstruction efforts were closely tied to autonomous driving applications. New College~\cite{newcollege} and NCLT~\cite{NCLT} introduced outdoor datasets incorporating GPS and LiDAR, establishing benchmarks for subsequent research. Semantic3D~\cite{hackel2017semantic3d} provided large-scale outdoor datasets with over 3–4 billion labeled points across 15 outdoor scenes. The KITTI dataset~\cite{geiger2012we}, leveraging LiDAR scans and images, became a fundamental benchmark for autonomous driving, later extended by SemanticKITTI~\cite{Behley2019SemanticKITTIAD}, which enriched it with fine-grained semantic segmentation annotations.  
To achieve precise ground truth, New College~\cite{newcollege} Dataset utilized Terrestrial Laser Scanning (TLS)~\cite{TLS} for centimeter-level accuracy, while Hilti-Oxford~\cite{HiltiOx} attained millimeter precision. ETH3D~\cite{ETH3D} leveraged high-resolution imagery for highly accurate ground truth data. Additionally, Tanks and Temples integrated Structure-from-Motion (SfM) and Multi-View Stereo (MVS) techniques to establish challenging reconstruction benchmarks~\cite{Vijayanarasimhan2017SfMNetLO, MVS}. SemanticKITTI~\cite{Behley2019SemanticKITTIAD} further advanced outdoor scene segmentation by providing semantically labeled LiDAR point clouds.  
Beyond autonomous driving, datasets have expanded to more diverse and complex environments. Complex Urban~\cite{ComplexUrban} and WoodScape~\cite{woodscape} introduced urban-scale diversity, while large-scale driving datasets such as Argoverse (2019)~\cite{Argoverse}, nuScenes (2019)~\cite{Caesar2019nuScenesAM}, and Waymo Open~\cite{waymo} provided extensive multimodal data for perception and localization tasks. More recently, ARKitScene~\cite{baruch2022arkitscenesdiverserealworlddataset} and Habitat-Matterport 3D~\cite{habitat} have integrated RGB and LiDAR data to support AR/VR and navigation applications. Notably, Waymo Open~\cite{waymo} remains one of the largest autonomous driving datasets, featuring high-quality camera and LiDAR data.  
Despite these advancements, heritage and cultural sites have been underrepresented in large-scale reconstruction efforts. The ArCH benchmark addressed this gap by introducing 17 annotated point cloud scenes of heritage and architectural cultural sites~\cite{matrone2020arch}. Similarly, the Gibson Environment provided 572 building scans for embodied agent simulation, although it lacked detailed semantic annotations~\cite{savva2017gibson}. More recent datasets, such as Toronto-3D and SensatUrban~\cite{hu2022sensaturbanlearningsemanticsurbanscale}, have enhanced city-scale mapping through detailed point cloud data. These developments underscore the growing need for versatile, high-quality, and culturally diverse datasets that can support a wide range of reconstruction, segmentation, and localization applications in both research and real-world deployments.  MatrixCity~\cite{matrixcity} offers a large-scale, high-quality synthetic city environment designed to advance research in city-scale neural rendering and related applications. However, its synthetic nature may not fully capture the complexities of real-world scenarios, potentially limiting its applicability in practical settings.
GauU Scene~\cite{gauu_scene} introduces a novel large-scale scene reconstruction dataset utilizing Gaussian Splatting, encompassing over 1.5 square kilometers with comprehensive RGB and LiDAR ground truth data. But the dataset's relatively homogeneous scenes, primarily focused on outdoor environments, may restrict its utility across diverse settings. Our goal is to introduce a larger-scale, higher-quality, and more diverse large-scale 3D scene dataset to support research on large-scale Gaussian Splatting.

\subsection{Large-Scale 3D Reconstruction Methods}

Large-scale 3D reconstruction methods have evolved from traditional Structure-from-Motion and Multi-View Stereo pipelines, such as Tanks and Temples \cite{knapitsch2017tanks} and COLMAP \cite{schonberger2016structure}, to modern learning-based approaches. While classical methods effectively recover camera poses and geometric structures, they suffer from memory limitations and blurring artifacts in large-scale scenes. To address these challenges, recent deep learning-based techniques, such as 3D Gaussian Splatting\cite{kerbl3Dgaussians}, Surface-Aligned Gaussian Splatting (SuGaR) \cite{sun2023sugar}, and Gaussian Opacity Fields (DOF)\cite{lee2024dof}, have demonstrated significant improvements in room-scale scene reconstruction. Moreover, the latest CityGaussian model exhibits state-of-the-art neural rendering capabilities, enabling the reconstruction of complex urban environments \cite{doe2023citygaussian}.  
Despite these advancements, most existing benchmarks rely on constrained or synthetic datasets, primarily due to the lack of high-resolution image data and the limited representation of culturally diverse scenes. This gap hinders the generalizability and scalability of 3D reconstruction methods. Our proposed dataset aims to address this limitation by providing a large-scale, high-fidelity benchmark that fosters advancements in detail-preserving 3D reconstruction.  
With improvements in point cloud data acquisition and deep learning architectures, large-scale point cloud datasets have become instrumental in application-driven research areas, including architectural reconstruction, semantic segmentation, and robotic navigation. These developments have significantly contributed to the evolution of end-to-end reconstruction models and data-driven training strategies.  
The table\ref{tab:merged_overview} provides an overview of existing open-source datasets, highlighting their primary features and applications. By comparing data acquisition methods, data types, dataset scale, and scene diversity, it is evident that the proposed Culture3D dataset offers significant advantages. With its combination of broad coverage and high fidelity, it provides a useful reference point for evaluating next-generation large-scale Gaussian splatting approaches.

\section{Dataset}
\label{sec:methodology}
This section covers data acquisition and preliminary reconstruction, leading to benchmarking and evaluation of 3D reconstruction methods. This analysis provides insights into algorithm performance across diverse scenes, detailed in the following subsections.

\begin{figure*}[h!]
    \centering
    \includegraphics[width=\linewidth]{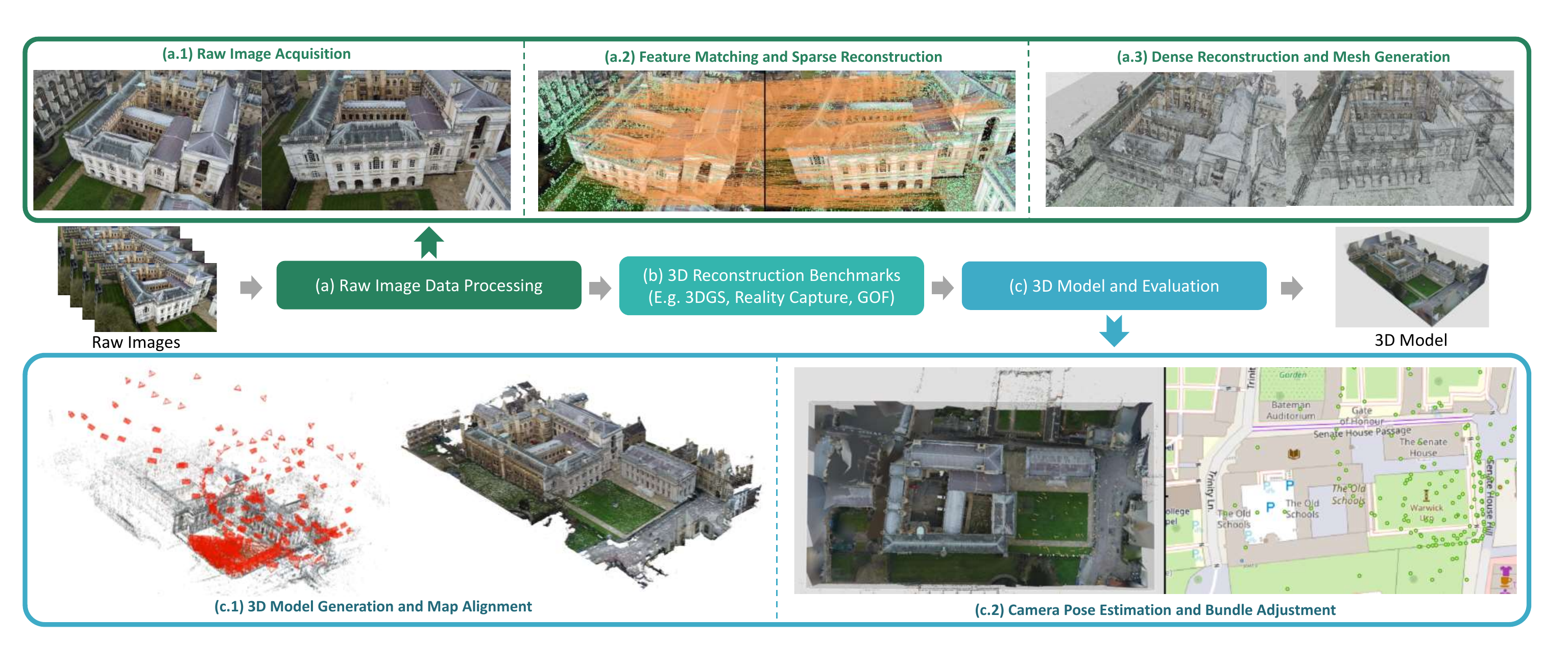}
    \caption{Main pipeline of our dataset. The workflow includes: (a) Raw image data processing (image acquisition, feature matching, sparse reconstruction, dense reconstruction, and mesh generation); (b) 3D Gaussian-Based Scene Rendering benchmarks; and (c) 3D model generation and evaluation (map alignment, camera pose estimation, and bundle adjustment).}
    \vspace{-3mm}
    \label{fig:dataset_structure}
\end{figure*}

\subsection{Dataset Collection}

 Starting with raw image data collection, our dataset CULTURE3D comprises 41,006 high-resolution images (48MP each) using a DJI Mini 3 drone equipped with a 1/1.3-inch CMOS sensor capable of 4K HDR video recording, ensuring both high-resolution stills and dynamic video data. The imaging system features an f/1.7 aperture, an ISO range of 100–3,200, electronic shutter speeds ranging from 2 to 1/8,000 s, and a maximum image resolution of 8064×6048 pixels. Coupled with a 3-axis mechanical gimbal for enhanced stabilization, the DJI Mini 3 achieves excellent performance under diverse lighting and weather conditions. The drone's mobility enabled extensive coverage with systematic flight paths designed for optimal image overlap around 15 degrees of angle. For indoor scenes, the same camera—mounted on a steady cloud-platform—was used to capture similar data sequences while minimizing pedestrian interference. Both indoor and outdoor scenes of cultural landmarks were acquired using controlled orbit and grid flight patterns, ensuring consistent camera parameter estimation and high-fidelity reconstructions.

\subsection{3D Reconstruction and Point Cloud Generation}

After getting all the high-resolution images, we used photogrammetry tools COLMAP and Reality Capture to produce both dense and sparse point cloud data stored in standard formats (.ply and .pcd). Reality Capture further refined these results, generating dense textured meshes and camera intrinsic and extrinsic parameters, which also enables downstreaming applications. The data in our dataset includes sparse reconstructions to aid in evaluations and further detailed analysis.

\subsection{3D Modeling and Asset Generation}
For further usage like virtual reality related applications, our dataset also provides multiple reconstructed 3D assets that were textured within Reality Capture. These assets are generated based on high-accuracy point cloud data therefore can support extensive applications in navigation, localization and ai-driven tasks.

\begin{figure*}[t]
    \centering    \includegraphics[width=\linewidth]{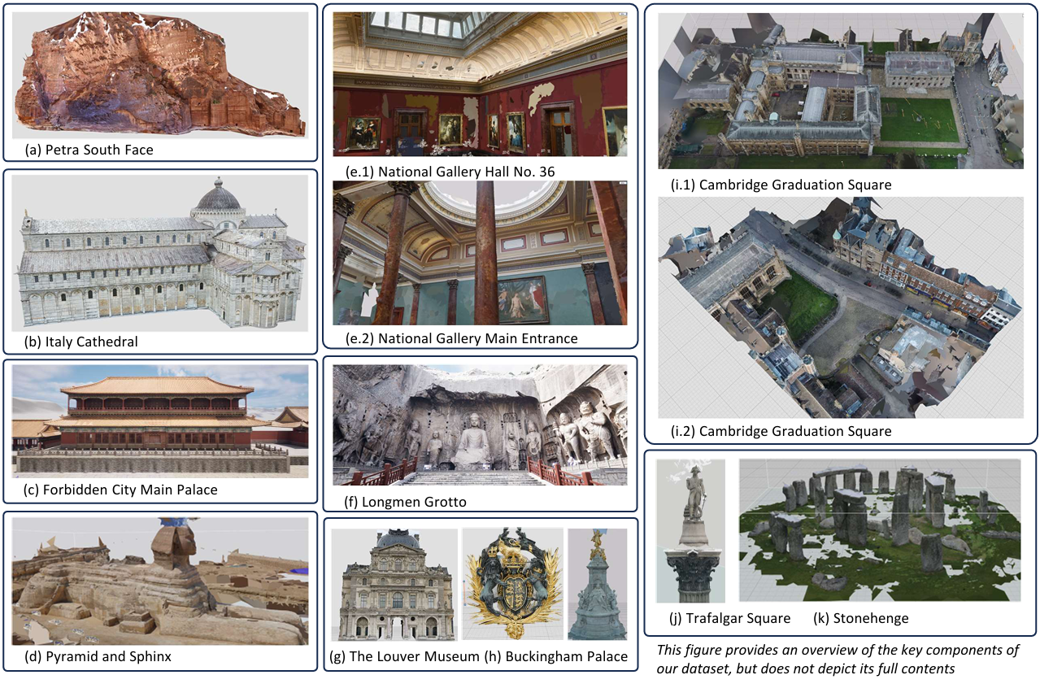}
    \vspace{-5pt}
    \caption{Overview of representative cultural heritage and urban environment datasets used in our benchmark.}
    \label{fig:overall}
\end{figure*}

\figurename~\ref{fig:overall} demonstrates key scenes from our dataset. (a) The \textit{Petra} dataset covers detailed reconstructions of natural stone formations around the Treasury. (b) The \textit{Leaning Tower of Pisa} dataset (labeled as “Italy Cathedral”) features the tower, cathedral, and surrounding area for structural analysis and VR tourism. (c) The \textit{Forbidden City} dataset captures intricate roof patterns and ornate carvings significant for heritage studies. (d) The \textit{Pyramids and Sphinx} dataset includes both aerial and ground-level imagery for detailed 3D modeling. (e.1-e.2) The \textit{National Art Gallery} dataset provides high-resolution interior and entrance views for artistic preservation. (f) The \textit{Longmen Grottoes} dataset emphasizes fine carvings and environmental detail. (g) The \textit{Louvre Museum} dataset offers extensive interior and exterior coverage for virtual reality and architectural modeling. (h) The \textit{Buckingham Palace} dataset highlights architectural features including the entrance gate and façade details. (i.1-i.2) The \textit{Cambridge Campus} dataset includes major buildings and pathways suitable for virtual tours. (j) The \textit{Trafalgar Square} dataset showcases iconic statues and surrounding architectural elements. Finally, (k) the \textit{Stonehenge} dataset focuses on the monument’s unique stone arrangement for archaeological research and VR applications.

\subsection{Discussion on the Limitations of CULTURE3D} 

Despite the high quality of our dataset, certain challenges remain. Minor calibration errors may arise due to drone movement, and slight photometric inconsistencies can occur as a result of environmental variations. Additionally, our dataset primarily focuses on static scenes, with efforts made to capture images in a manner that minimizes the impact of dynamic objects and extreme lighting conditions. To further enhance dataset reliability, we aim to mitigate errors caused by environmental fluctuations and provide insights for future dataset expansions and methodological improvements. Additional dataset visualizations can be found in the supplementary materials.
\section{Experiments and Baselines}
\label{sec:baseline}

\begin{table*}[t]
 
  \label{tab:scene_comparison}
  \scriptsize
  \centering
  \begin{tabular}{llccccc}
    \toprule
    \textbf{Method} & \textbf{Metric} & \textbf{Cambridge Graduation Square} & \textbf{Trinity St East} & \textbf{Petra Treasury Face} & \textbf{Gallery Hall No.36} & \textbf{Pyramid} \\
    \midrule
    \multirow{3}{*}{3DGS} 
    & SSIM $\uparrow$  & 0.2861 & 0.6268 & 0.4901 & 0.3934 & 0.4478 \\
    & PSNR $\uparrow$ & 13.3889 & 17.7258 & 17.0699 & 11.6312 & \textbf{18.2355} \\
    & LPIPS $\downarrow$ & 0.5872 & 0.3812 & \textbf{0.4703} & 0.5669 & 0.5631 \\
    & Time (hrs)     & 0.6  & 0.52  & 0.95  & 0.48  & 0.75   \\

    \midrule
    \multirow{3}{*}{SuGaR} 
    & SSIM $\uparrow$ & OOM & 0.6048 & 0.4758 & 0.4129 & 0.4205 \\
    & PSNR $\uparrow$ & OOM & 18.1793 & 17.0270 & 12.6912 & 17.4085 \\
    & LPIPS $\downarrow$ & OOM & 0.4428 & 0.4784 & 0.5823 & \textbf{0.5524} \\
    & Time (hrs) & OOM & 8.17 & 23.95 & 12.42 & 17.75 \\

    \midrule
    \multirow{3}{*}{Wild Gaussian} 
    & SSIM $\uparrow$ & OOM & 0.6262 & 0.6074 & 0.2069 & 0.4206 \\
    & PSNR $\uparrow$ & OOM & 18.3636 & 19.8103 & 8.3311 & 14.8998 \\
    & LPIPS $\downarrow$ & OOM & 0.3736 & 0.4880 & 0.8256 & 0.8193 \\
    & Time (hrs) & OOM & 2.98 & 23.83 & 13.80 & 22.42 \\
    \midrule
    \multirow{3}{*}{GOF} 
    & SSIM $\uparrow$ & 0.2735 & \textbf{0.7167}  & 0.5379 & 0.4233 & FAIL \\
    & PSNR $\uparrow$ & 15.3888 & 19.3954 & 19.8737 & \textbf{18.3617} & FAIL  \\
    & LPIPS $\downarrow$ & \textbf{0.4824} & \textbf{0.0260} & 0.4749 & 0.5764 & FAIL \\
    & Time (hrs)  & 18.50 & 17.20 & 15.43 & 9.97 & FAIL \\
    \midrule
    \multirow{3}{*}{HoGS} 
    & SSIM  $\uparrow$ & OOM & 0.5737 & 0.4190 & 0.4011 & FAIL \\
    & PSNR  $\uparrow$ & OOM & 19.6982 & 18.0278 & 12.6378 & FAIL \\
    & LPIPS $\downarrow$ & OOM & 0.4091 & 0.4918 & \textbf{0.4682} & FAIL \\
    & Time (hrs) & OOM & 4.34 & 6.82 & 7.67 & FAIL \\
    \midrule
    \multirow{3}{*}{City GS} 
    & SSIM  $\uparrow$ & \textbf{0.6129} & 0.6526 & \textbf{0.6998} & \textbf{0.6245} & \textbf{0.5647} \\
    & PSNR  $\uparrow$ & \textbf{16.6261} & \textbf{21.68} & \textbf{21.86} & 14.3655 & 17.4918 \\
    & LPIPS $\downarrow$& 0.7855 & 0.6397 & 0.6354 & 0.8001 & 0.7402 \\
    & Time (hrs) & 7.69 & 5.84 & 6.59 & 8.56 & 12.16 \\
    \bottomrule
  \end{tabular}
   \caption{Comparison of 3D scene reconstruction methods across datasets. Arrows indicate desired metric performance ($\uparrow$ higher is better, $\downarrow$ lower is better). Bold numbers highlight the best-performing results. "OOM" denotes out-of-memory errors; "FAIL" indicates reconstruction failures.}
    \vspace{-3mm}
\end{table*}

\begin{figure*}[h]
  \centering
  \scriptsize
  \setlength{\tabcolsep}{2mm}
  \renewcommand{\arraystretch}{1.5}
  \begin{tabular}{>{\centering\arraybackslash}m{2cm} >{\centering\arraybackslash}m{4cm} >{\centering\arraybackslash}m{4cm} >{\centering\arraybackslash}m{4cm}}
  \midrule
    \textbf{Benchmark} & \textbf{Cambridge Graduation Square} & \textbf{Cambridge Trinity St East} & \textbf{Petra Treasury Face} \\
    \midrule
    \textbf{Ground Truth} & 
      \includegraphics[width=0.2\textwidth]{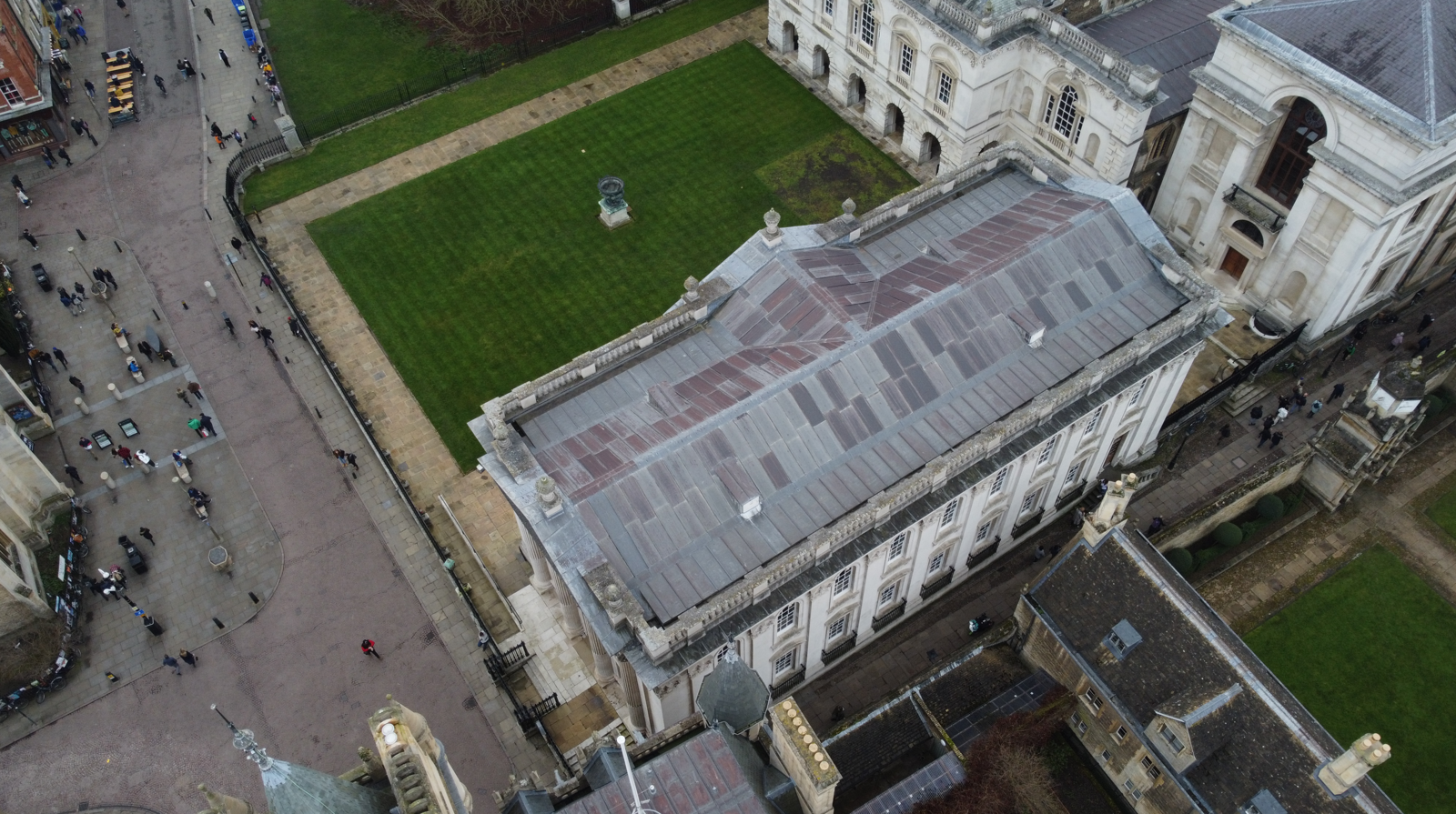} & 
      \includegraphics[width=0.2\textwidth]{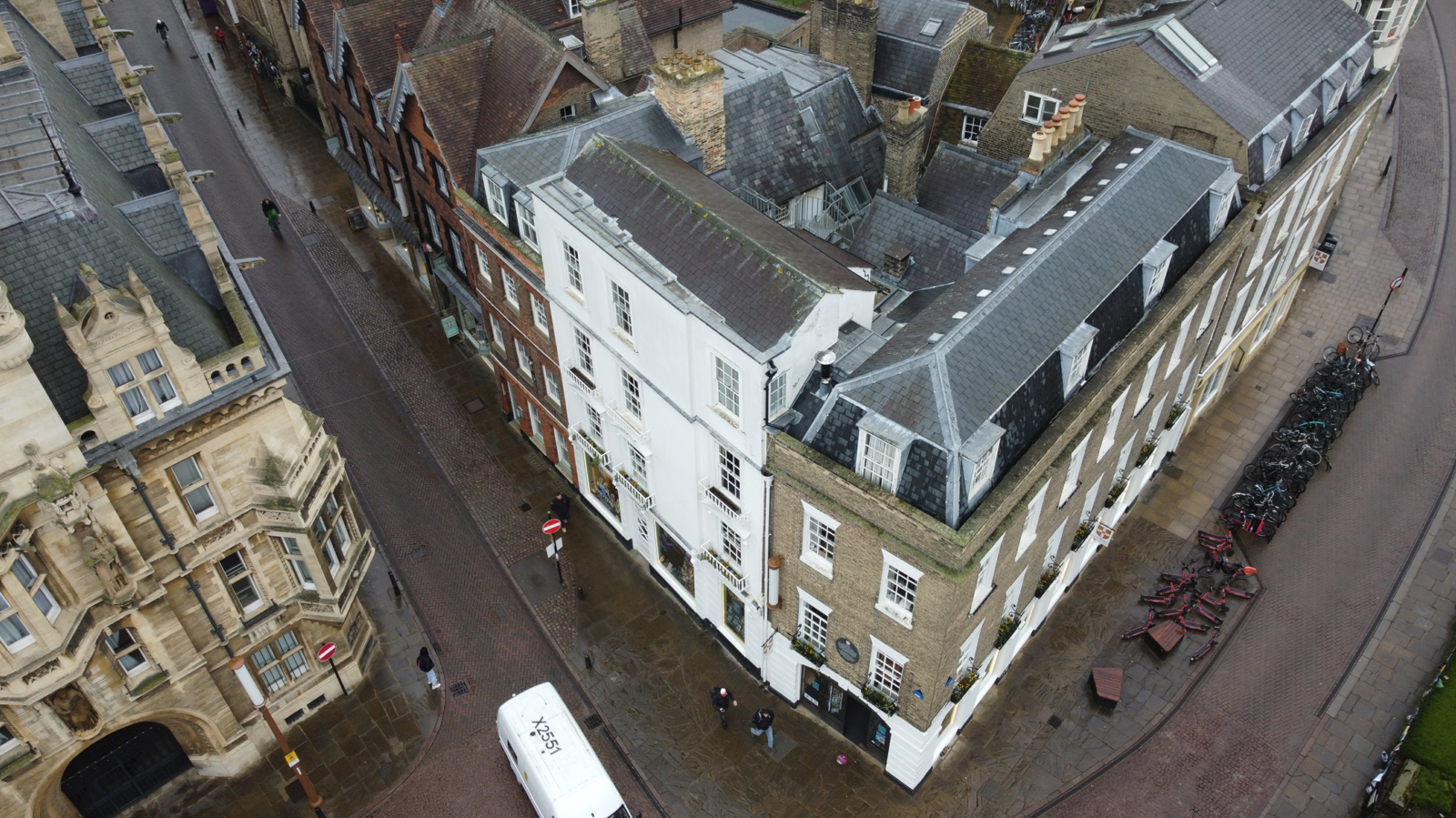} & 
      \includegraphics[width=0.18\textwidth]{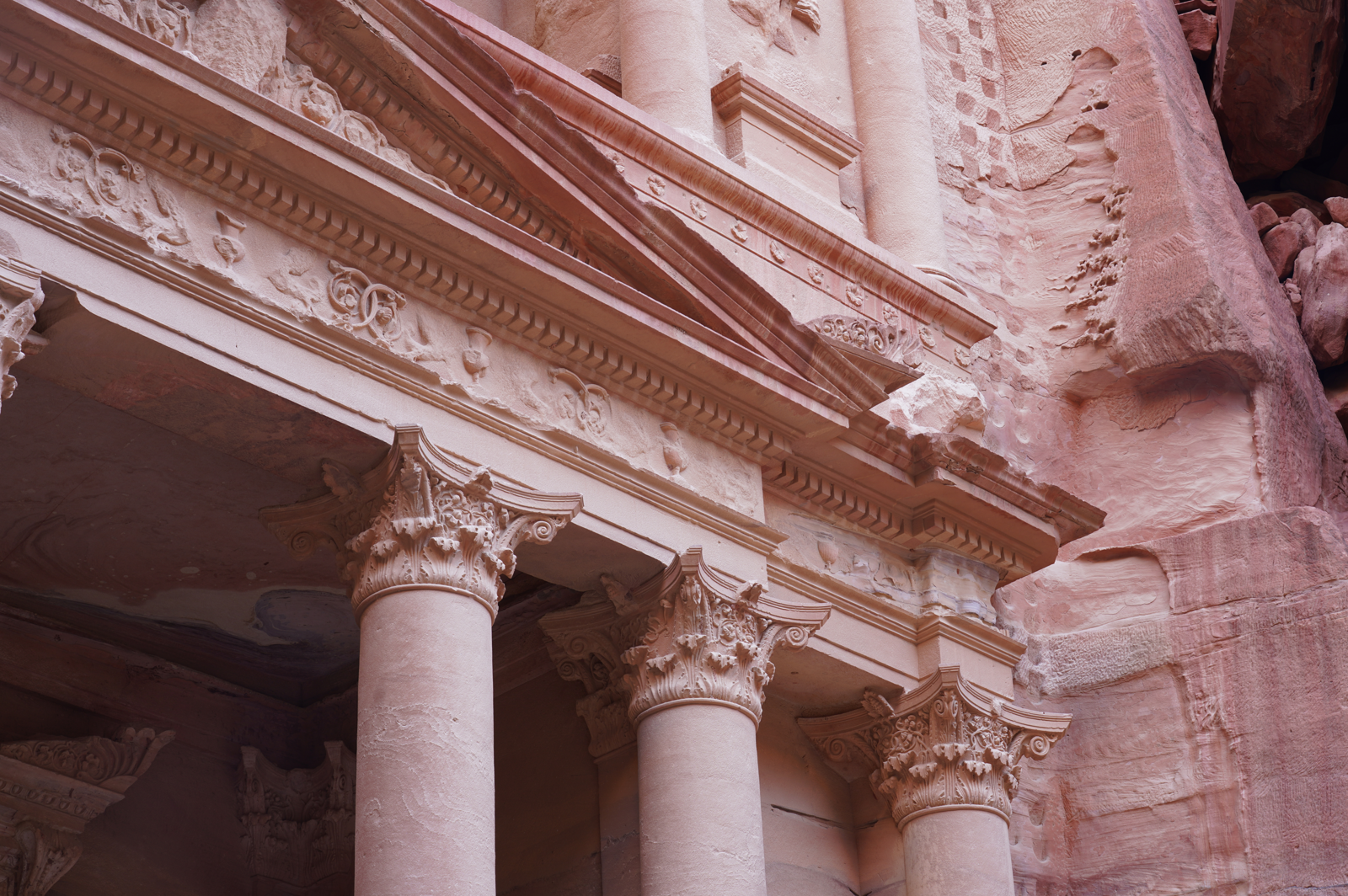} \\
      \textbf{Reality Capture} & 
      \includegraphics[width=0.2\textwidth]{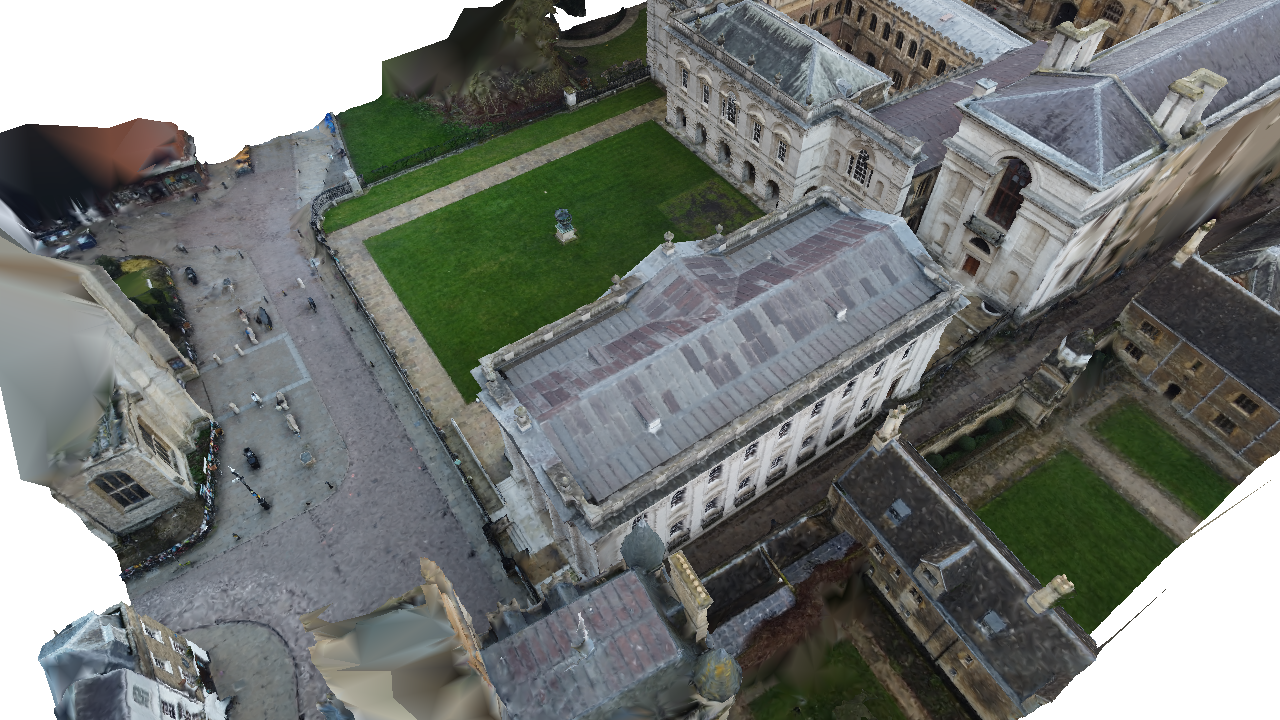} & 
      \includegraphics[width=0.2\textwidth]{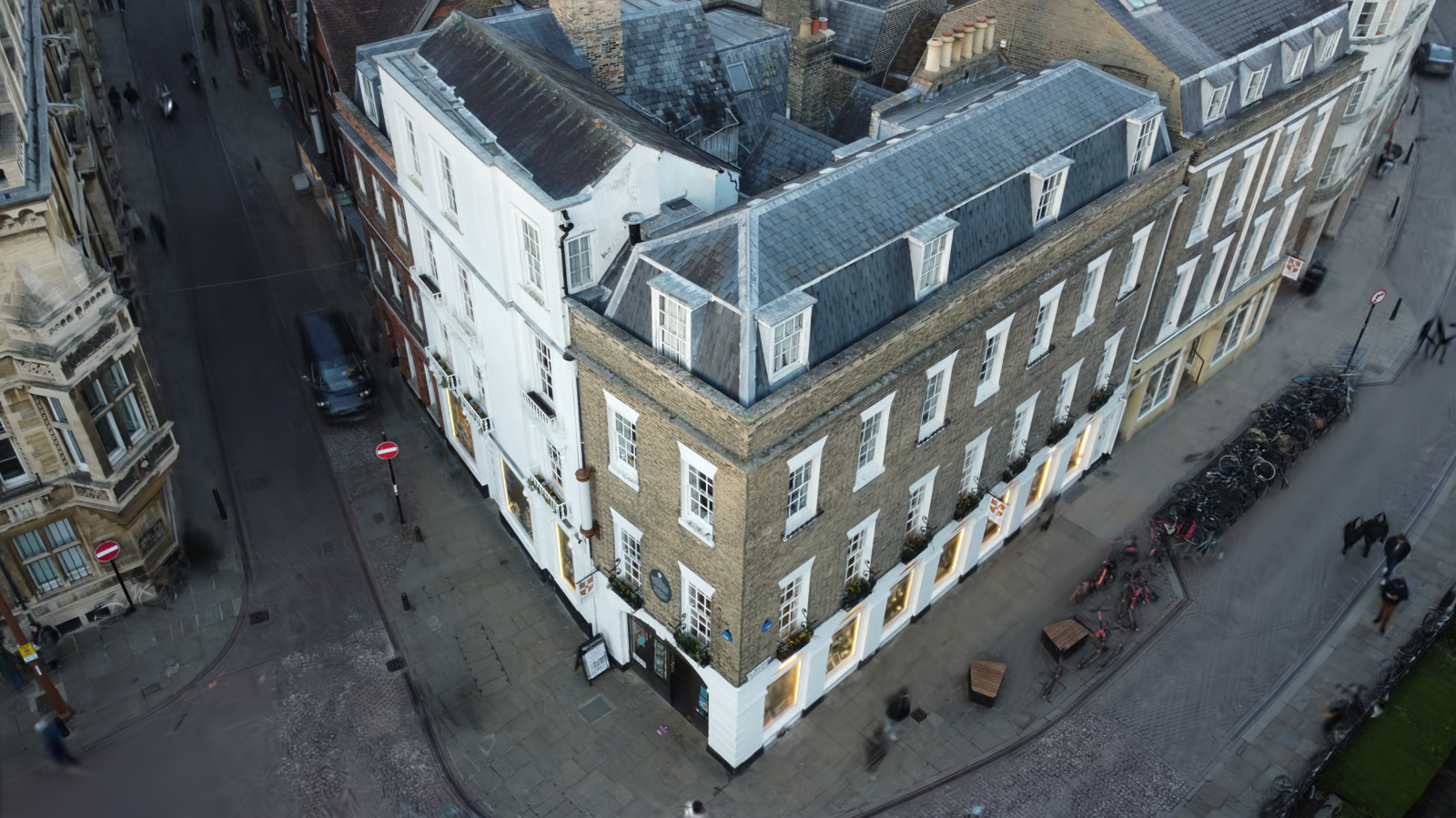} & 
      \includegraphics[width=0.18\textwidth]{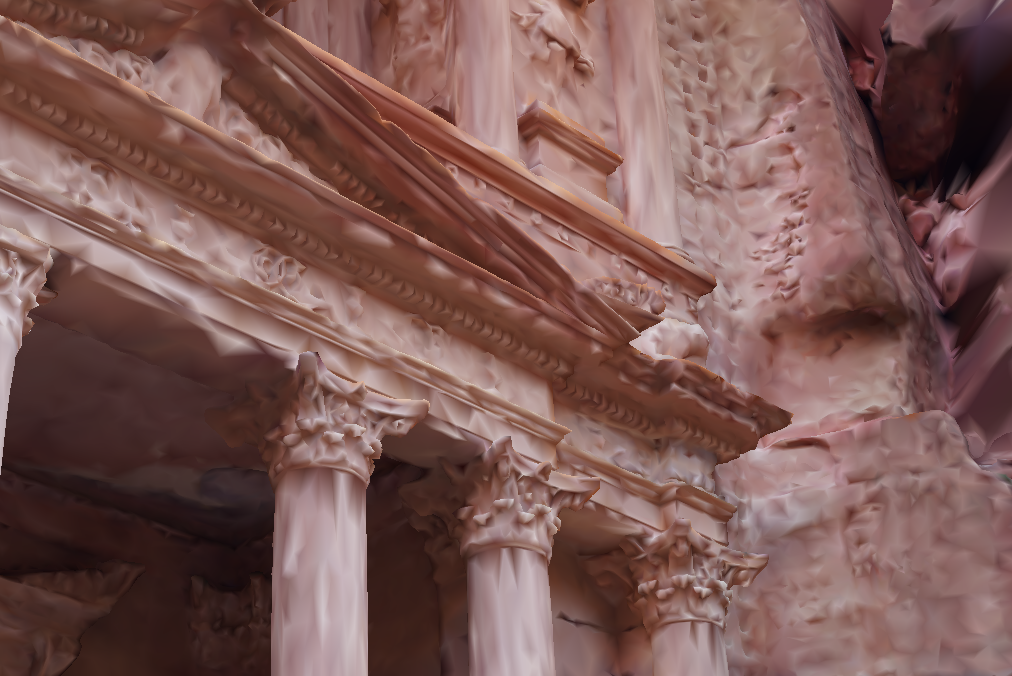} \\
    \textbf{3DGS} & 
      \includegraphics[width=0.2\textwidth]{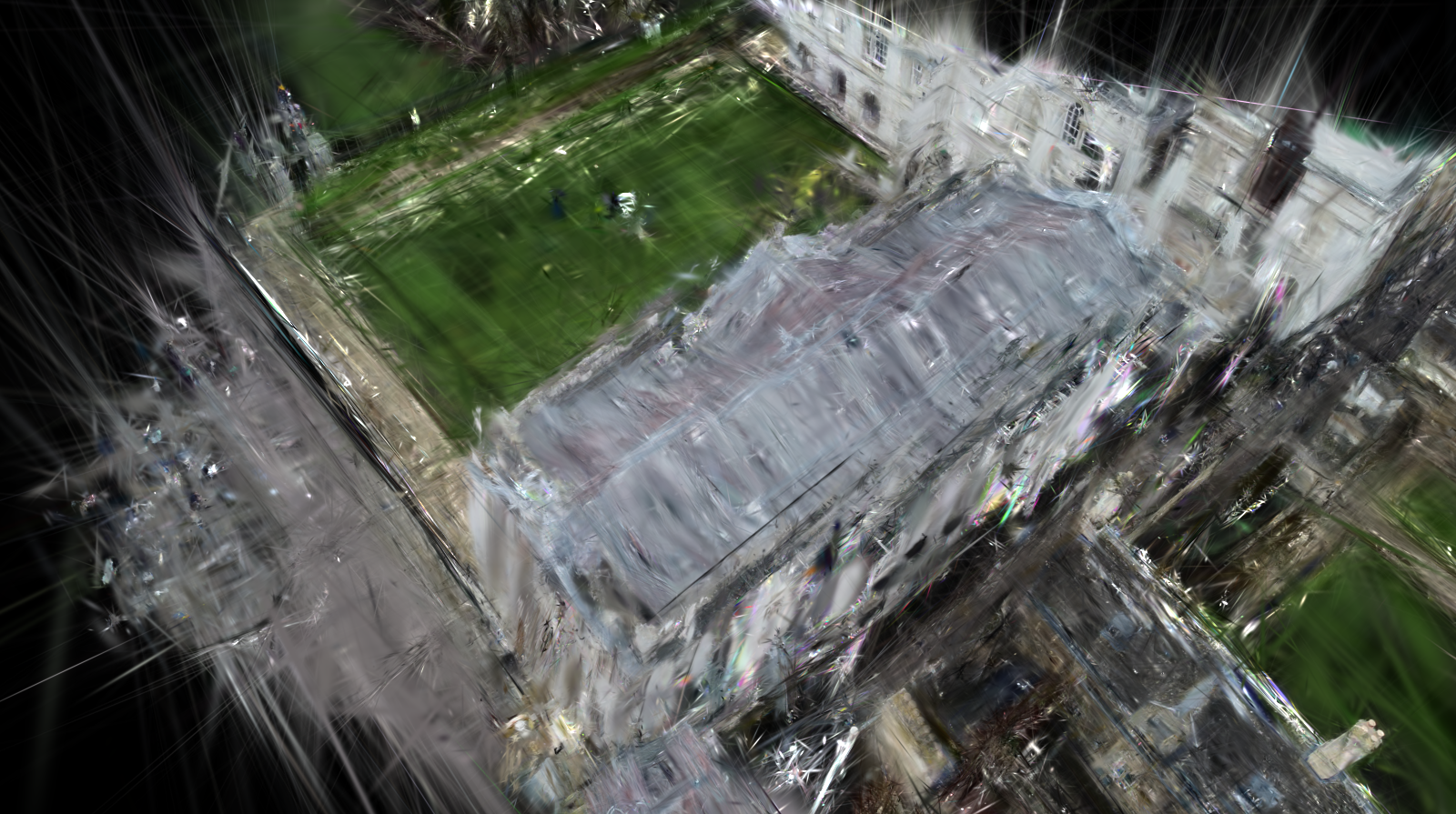} & 
      \includegraphics[width=0.2\textwidth]{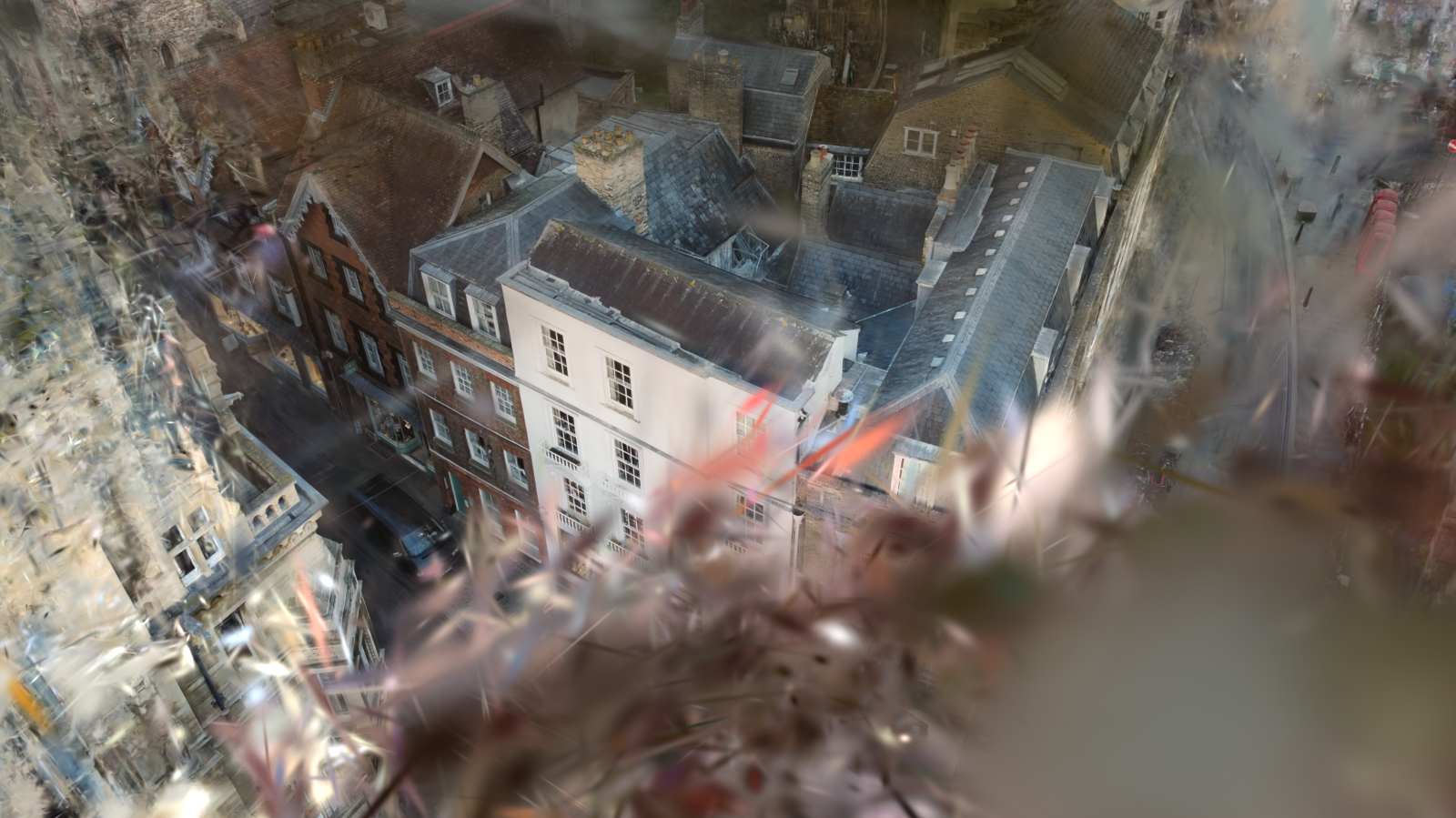} & 
      \includegraphics[width=0.18\textwidth]{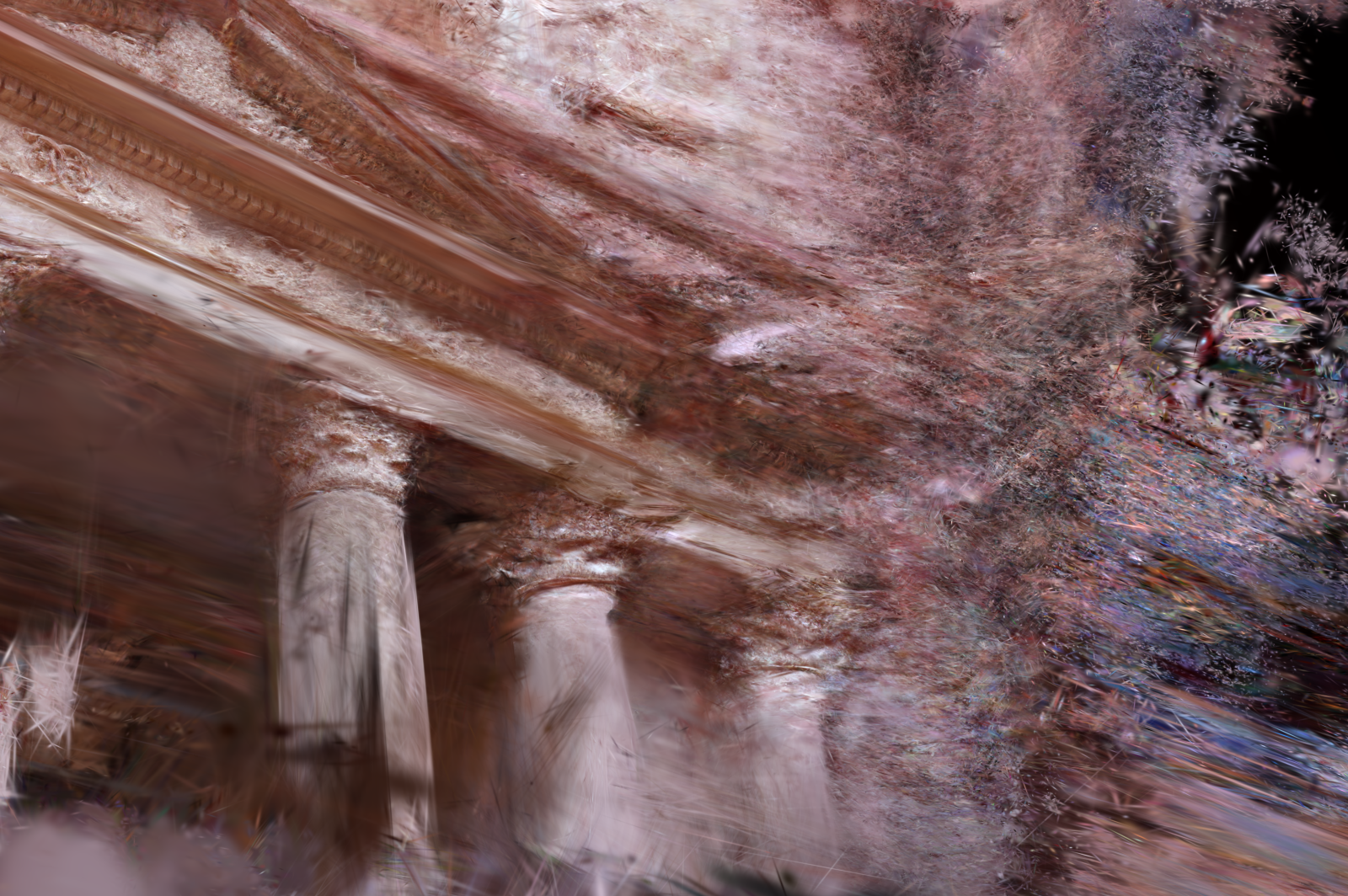} \\
    \textbf{SuGaR} & 
      \textbf{Out Of Memory} & 
      \includegraphics[width=0.2\textwidth]{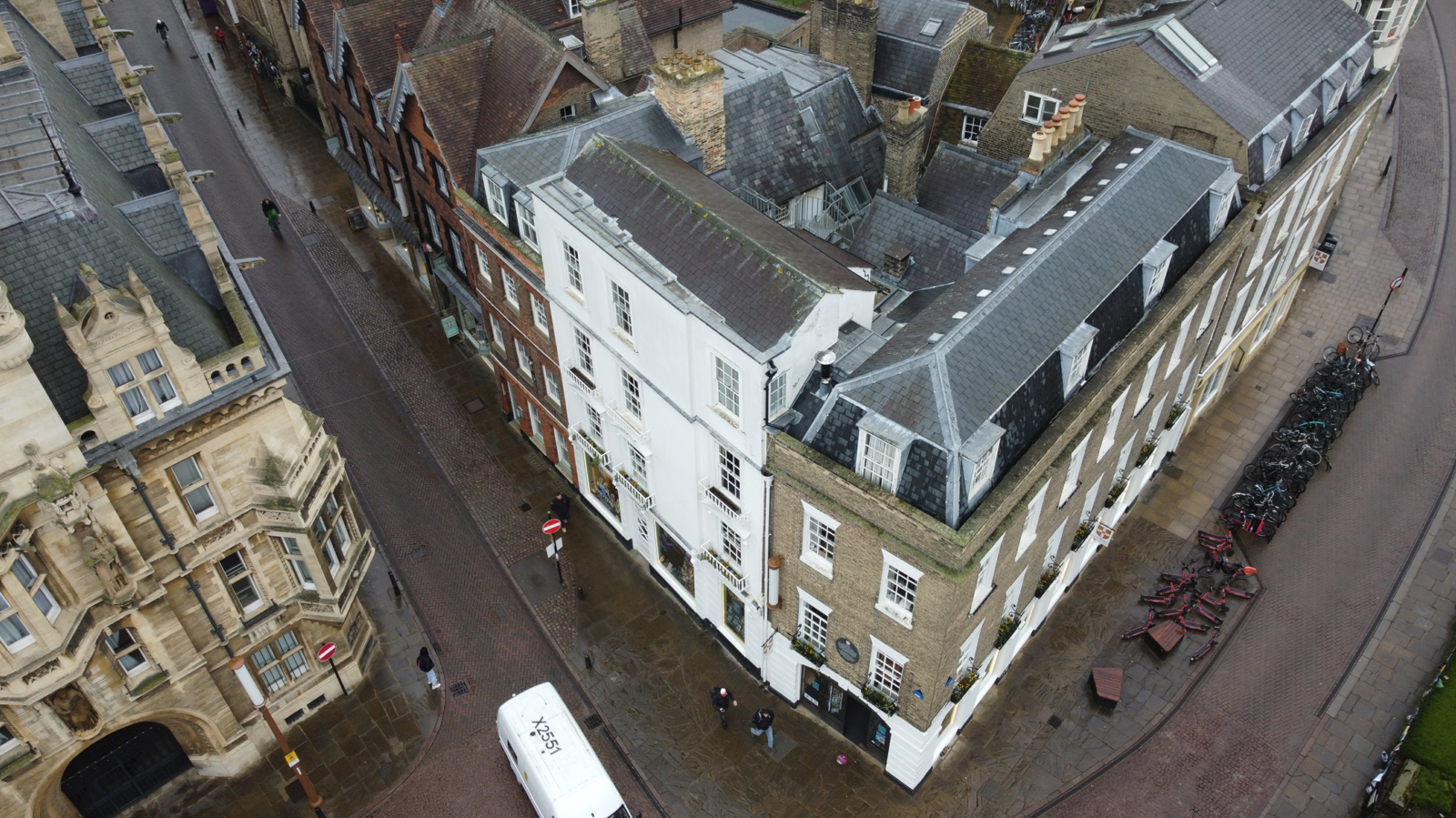} & 
      \includegraphics[width=0.18\textwidth]{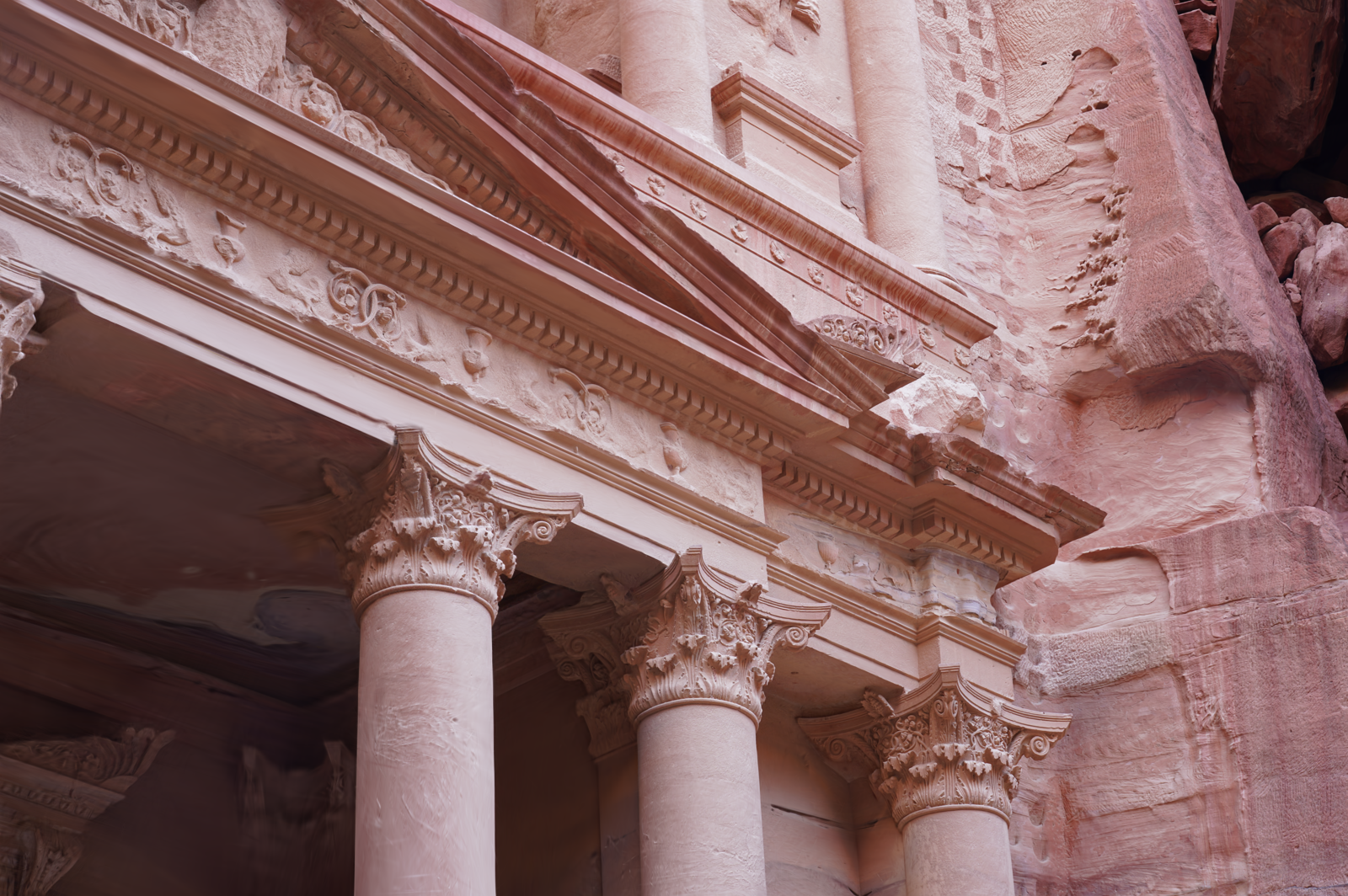} \\
    \textbf{Wild Gaussian} & 
      \textbf{Out Of Memory} & 
      \includegraphics[width=0.2\textwidth]{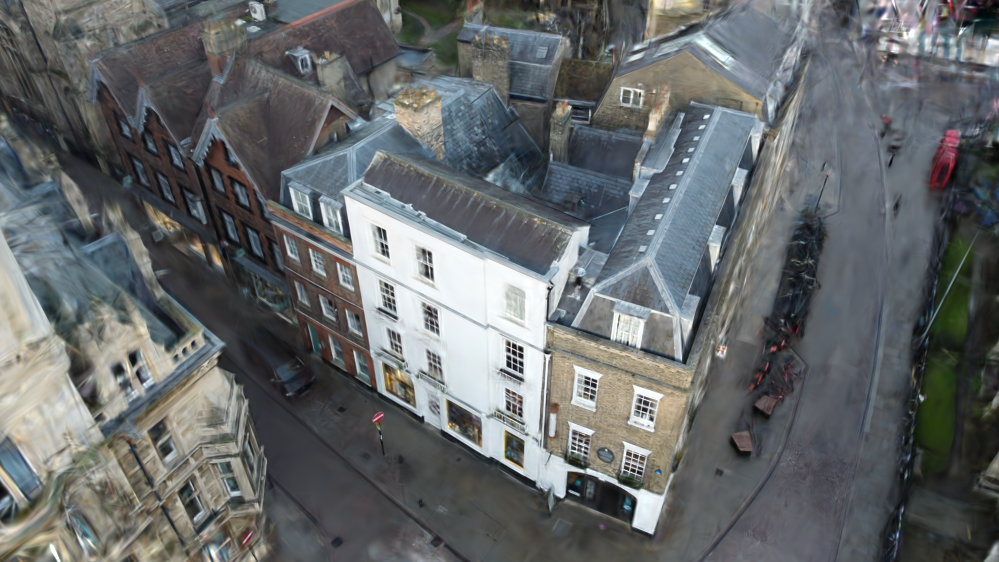} & 
      \includegraphics[width=0.18\textwidth]{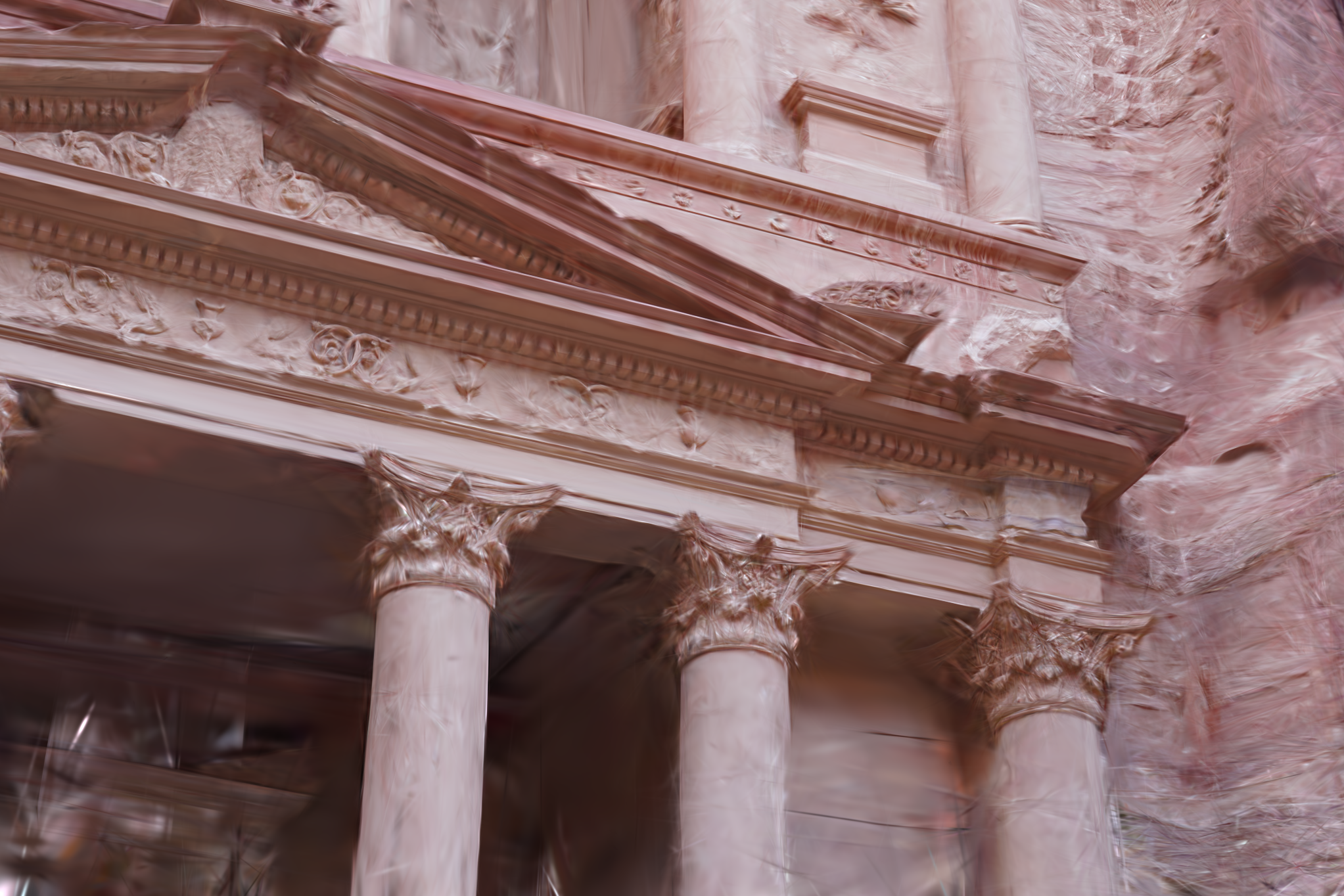} \\
      \textbf{GOF} & 
      \includegraphics[width=0.2\textwidth]{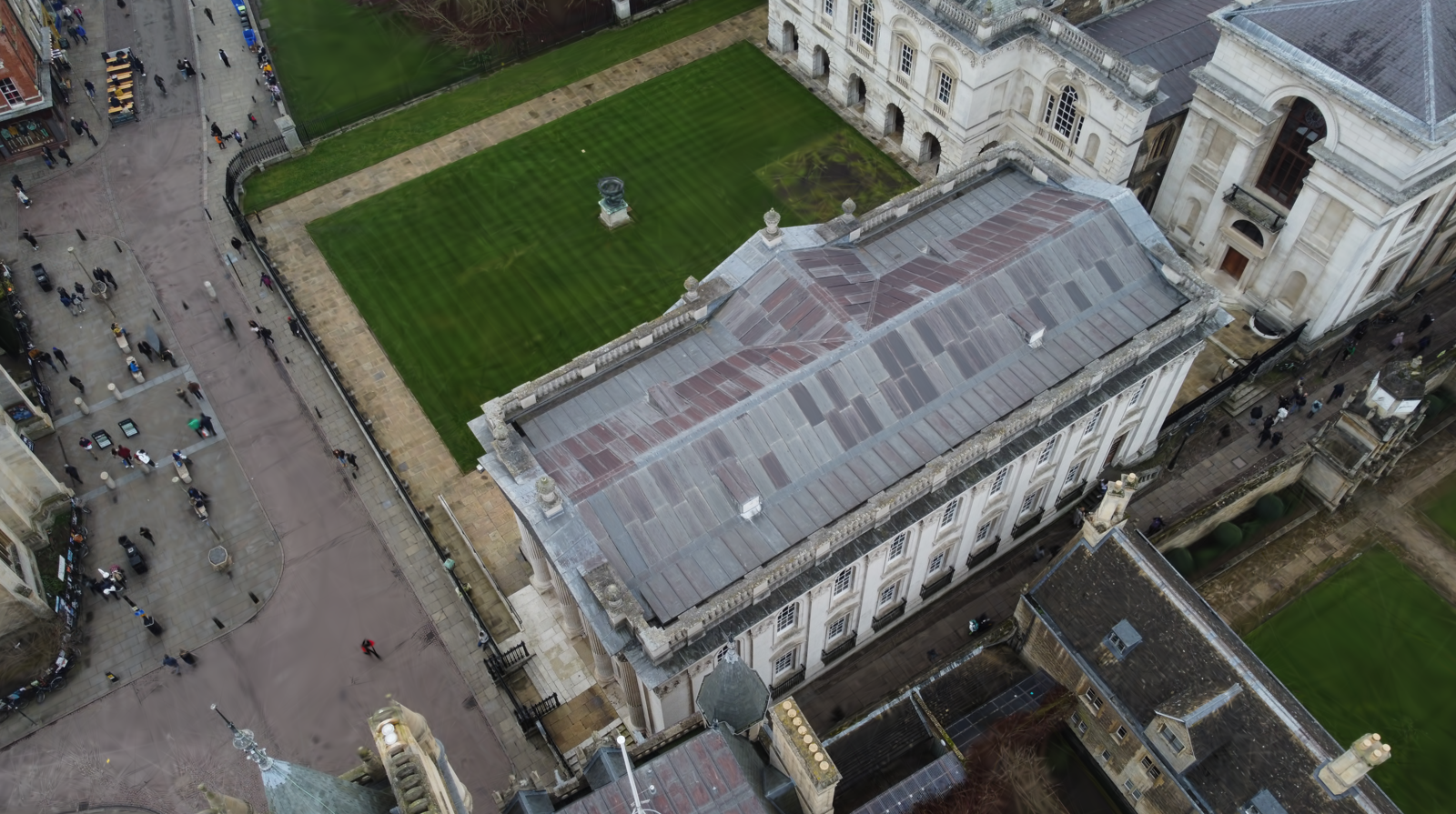} & 
      \includegraphics[width=0.2\textwidth]{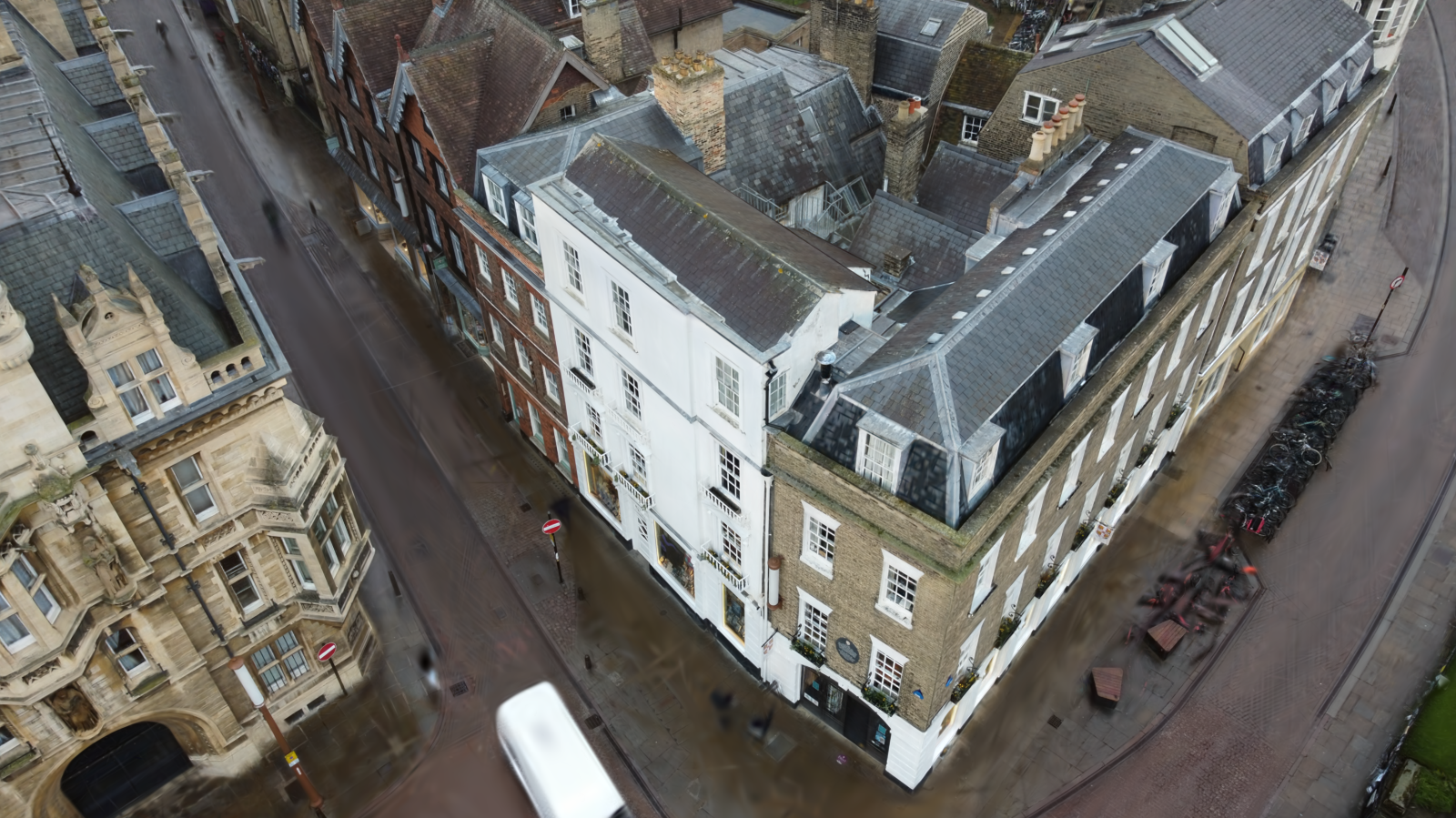} & 
      \includegraphics[width=0.18\textwidth]{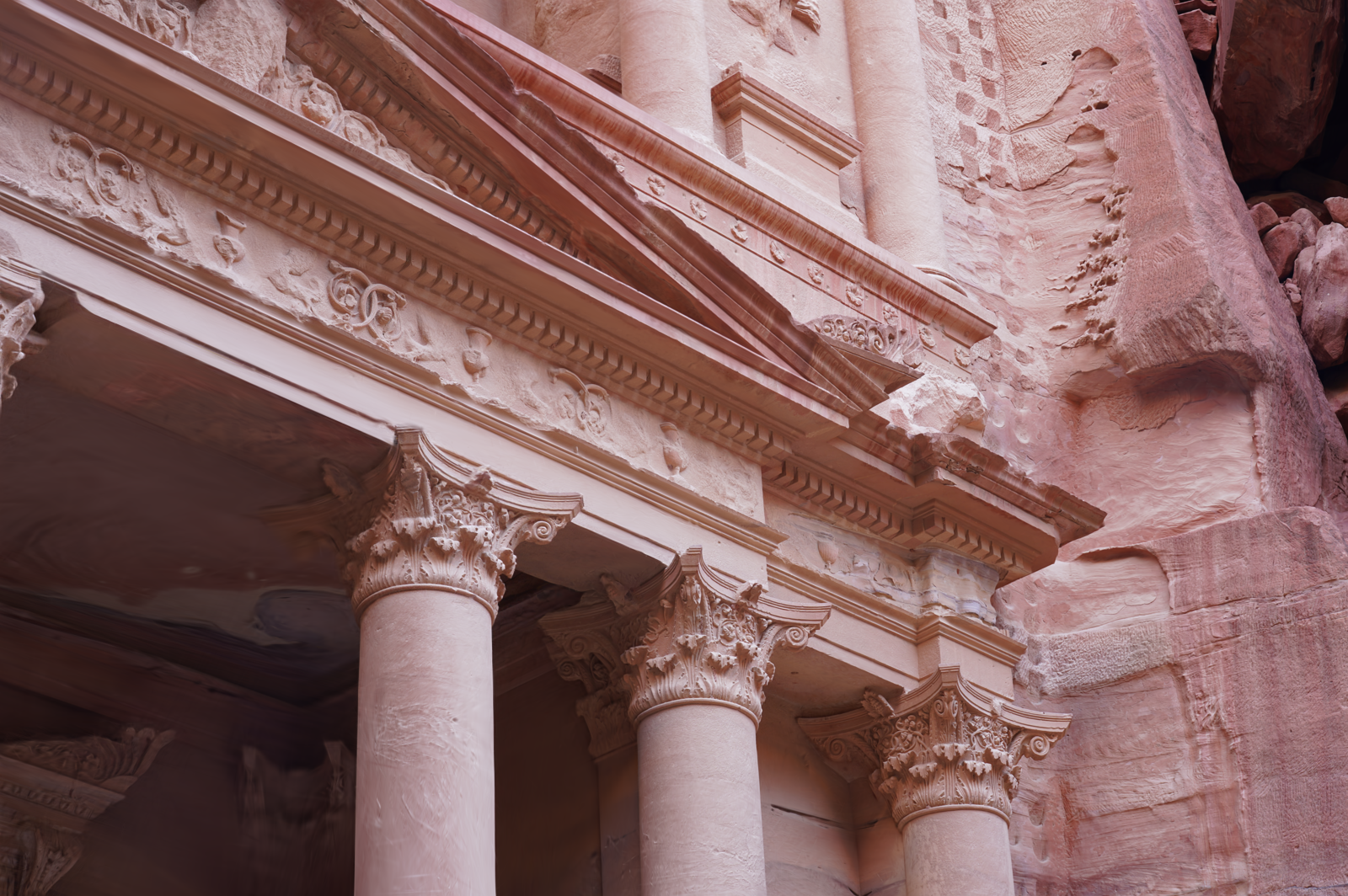} \\
      \midrule
  \end{tabular}
    \caption{Qualitative comparison of reconstructed 3D scenes across multiple methods. The rows represent different reconstruction methods, while columns show three benchmark scenes.}
    \label{fig:scene_comparison}
    \vspace{-5mm}
\end{figure*}

We evaluate multiple state-of-the-art large-scale Gaussian-based scene rendering methods on the CULTURE3D dataset to assess their reconstruction performance in large-scale scenes with high-resolution details.
Specifically, we compare the reconstruction results of four representative methods, including a neural point-based radiance field approach (3D Gaussian Splatting, 3DGS)~\cite{kerbl3Dgaussians}, a well-established photogrammetry-based method (RealityCapture), and three recent neural methods—Surface-Aligned Gaussian Splatting (SuGaR)~\cite{sun2023sugar}, Gaussian Opacity Fields (GOF)~\cite{lee2024dof} and in-the-wild Gaussian Splatting (Wild Gaussian\cite{kulhanek2024wildgaussians}). These methods span both traditional techniques and state-of-the-art neural-based approaches, providing a comprehensive analysis of performance in large-scale 3D environments.

\subsection{Benchmark Evaluation Pipeline}

To ensure fair evaluation across neural-based reconstruction methods, our benchmarking uses a consistent pipeline, unified data structure and standardized ground truth to provide objective evaluation results.

All methods use identical inputs—high-resolution images and a COLMAP-generated point cloud ground truth. We apply both recent neural approaches (from the past two years) and established methods like 3DGS, enabling direct comparison of rendered outputs (Figure~\ref{fig:scene_comparison}) to demonstrate consistency. Finally, each reconstruction is quantitatively evaluated against the same ground truth images to ensure fair comparison.

\subsection{Experimental Settings}
The key experimental configurations for our reconstruction methods are listed in Table~\ref{tab:exp_settings}, covering training and refinement iterations, position and feature learning rates, loss weights, and checkpoint evaluation parameters. This standardized setup ensures consistent, fair comparisons across methods. Detailed settings will be provided on the official dataset page.

\begin{table}[htbp]
\centering
\scriptsize

\begin{tabular}{lll}
\hline
\textbf{Parameter Group} & \textbf{Parameter} & \textbf{Setting} \\
\hline
\multirow{4}{*}{Hardware} 
  & GPUs Used                & 8 × NVIDIA RTX A6000 \\
  & GPU Memory               & 48 GB per GPU \\
  & CUDA Version             & 12.1 (nvcc) \\
  & Driver Version           & 550.127.05 \\[0.5ex]
\hline
\multirow{1}{*}{Software Environment} 
  & OS                        & Ubuntu 20.04\\
\hline
\multirow{2}{*}{Training / Refinement} 
  & Total Iterations          & 30,000\\
  & Refinement Iterations     & 15,000 (Sugar) \\[0.5ex]
\hline
\multirow{4}{*}{Learning Rates} 
  & Position LR (Initial)     & 0.00016 \\
  & Position LR (Final)       & 0.0000016 \\
  & Feature LR                & 0.0025 \\
  & Appearance Network LR     & 0.001 \\[0.5ex]
\hline
Loss Weighting 
  & $\lambda_{dssim}$         & 0.2 \\[0.5ex]
\hline
\end{tabular}
\caption{Key Experimental Settings.}
\vspace{-4mm}
\label{tab:exp_settings}
\end{table}

\subsection{Evaluation Metrics}
We evaluate the quality of the reconstructions using three metrics: Peak Signal-to-Noise Ratio (PSNR), Structural Similarity Index Measure (SSIM)~\cite{1284395}, and Learned Perceptual Image Patch Similarity (LPIPS)~\cite{8578166}. PSNR is used to measure pixel-level fidelity via the mean squared error, whereas SSIM assesses local similarity in contrast, luminance and structure between images.

LPIPS captures high-level perceptual differences by comparing deep feature representations, thereby addressing image quality aspects that align with human visual perception. This metric is particularly useful in detecting subtle structural and textural discrepancies that traditional pixel-based measures may overlook.

In summary, combining PSNR, SSIM, and LPIPS provides a balanced evaluation of both numerical accuracy and perceptual quality, ensuring that our assessments reflect objective measurements as well as subjective visual similarity.

\subsection{Benchmark Results and Failure Cases Analysis}

These benchmark results highlight the challenges posed by CULTURE3D’s large-scale, high-detail scenes—even top-tier methods encountered difficulties.

Applying traditional 3DGS yields moderate SSIM values (0.2861–0.6268), indicating reasonable structural similarity. PSNR ranges from roughly 13.4 dB to 18.2 dB, reflecting moderate pixel fidelity. Figure~\ref{fig:scene_comparison} highlights structural errors consistent with lower PSNR compared to other models. Despite moderate results, a clear performance gap with other datasets underscores existing methods' limitations in detailed, large-scale cultural heritage scenes.

The SuGaR method achieves slightly improved SSIM and PSNR compared to standard 3DGS by aligning and regularizing surfaces. Although SuGaR outperforms 3DGS, it occasionally faces Out-of-Memory (OOM) issues due to high computational demands, especially in complex scenes.

Wild Gaussian, tested for handling lighting variations and subtle real-world changes, improves dynamic reconstruction and LPIPS scores but struggles with our dataset. Particularly, it faces challenges with richly detailed, large-scale scenes, encountering OOM errors due to computational demands. Despite limitations, Figure~\ref{fig:scene_comparison} shows it captures fine details effectively, notably achieving the highest SSIM (0.6074) for Petra.

In architectural evaluation, Gaussian Opacity Fields (GOF) maintains structural accuracy using a continuous opacity field for surface extraction, achieving high SSIM and PSNR among earlier methods. However, GOF struggles with intricate details and pixel similarity in the Pyramid dataset, marking a critical area for future improvement.

We also evaluated recent benchmarks—HoGS and City GS—which outperform previous methods. HoGS is faster due to hierarchical Gaussian organization but performs below GOF and crashes (OOM) on Cambridge and Pyramid datasets. City GS achieves the highest scores across nearly all scenes, demonstrating excellent scalability and robustness for large, complex structures. Its optimized management of extensive Gaussian primitives prevents OOM errors, establishing it as the state-of-the-art for large-scale cultural heritage rendering.

Overall, these benchmarks highlight challenges posed by CULTURE3D's high-detail, large-scale scenes. Even state-of-the-art methods face significant computational scalability limits. Moreover, outdoor environments with variable lighting or dynamic elements remain challenging, particularly for methods initially designed for static scenes.

\subsection{Dataset Significance and Applications}

These benchmark results highlight the challenges posed by CULTURE3D’s large-scale, high-detail scenes—even top-tier methods encountered scalability issues (memory or incomplete reconstructions). Even though all the evaluation results reached a certain level of accuracy, there's still a huge gap compared to other datasets' testing results, showing that current methods have disadvantages in dealing with our cultural heritage. Our dataset incorporates diverse scenes, including dynamic objects, thin structures, large-scale scenes, and rich details of cultural heritage. All these features not only reveal the limitations of current models but also provide a more realistic and challenging evaluation standard. When using the CULTURE3D dataset, researchers can gain deeper insight into reconstruction methods that are robust, versatile, and applicable to real-world complexities.

\section{Conclusion and Future Work}
\label{sec:conclusion}

Our dataset, CULTURE3D, focuses on large-scale and diverse cultural heritage scenes and serves as a benchmark for multiple Gaussian-based scene rendering methods. Through a comprehensive analysis of these reconstruction results, we identify key challenges and limitations in handling large-scale, highly detailed scenes. While methods such as 3DGS, SuGaR, GOF, and Wild Gaussian achieve reasonable accuracy, scalability and computational complexity remain significant hurdles. Particularly, large and structurally intricate datasets, such as Cambridge Uni main buildings and Egyptian Pyramids of Giza, frequently encounter Out-of-Memory(OOM) errors and incomplete reconstructions. These methods also struggle with reconstructing dynamic objects, thin structures, and scenes with complex decorative patterns.
Our findings underscore significant opportunities for improving large-scale 3D reconstruction models, particularly in scalability and memory efficiency. Future research should prioritize overcoming challenges associated with complex architectural structures and intricate natural details while ensuring color fidelity and high-resolution texture preservation. Moreover, the integration of neural networks with traditional geometric techniques holds great potential for enhancing real-world detail capture, paving the way for more robust, efficient, and precise reconstruction solutions for practical applications.




\newpage
\begin{thebibliography}{54}
\providecommand{\natexlab}[1]{#1}
\providecommand{\url}[1]{\texttt{#1}}
\expandafter\ifx\csname urlstyle\endcsname\relax
  \providecommand{\doi}[1]{doi: #1}\else
  \providecommand{\doi}{doi: \begingroup \urlstyle{rm}\Url}\fi

\bibitem[A. et~al.(2017)A., J., Q.-Y., and V.]{Tanks_and_Temples}
Knapitsch A., Park J., Zhou Q.-Y., and Koltun V.
\newblock Tanks and temples: Benchmarking large-scale scene reconstruction.
\newblock \emph{ACM Transactions on Graphics (TOG)}, 36\penalty0 (4):\penalty0 1--13, 2017.

\bibitem[Armeni et~al.(2016)Armeni, Sener, Zamir, Jiang, Brilakis, Fischer, and Savarese]{armeni20163d}
Iro Armeni, Ozan Sener, Amir~R Zamir, Helen Jiang, Ioannis Brilakis, Martin Fischer, and Silvio Savarese.
\newblock 3d semantic parsing of large-scale indoor spaces.
\newblock In \emph{Proceedings of the IEEE Conference on Computer Vision and Pattern Recognition (CVPR)}, 2016.

\bibitem[Baruch et~al.(2022)Baruch, Chen, Dehghan, Dimry, Feigin, Fu, Gebauer, Joffe, Kurz, Schwartz, and Shulman]{baruch2022arkitscenesdiverserealworlddataset}
Gilad Baruch, Zhuoyuan Chen, Afshin Dehghan, Tal Dimry, Yuri Feigin, Peter Fu, Thomas Gebauer, Brandon Joffe, Daniel Kurz, Arik Schwartz, and Elad Shulman.
\newblock Arkitscenes: A diverse real-world dataset for 3d indoor scene understanding using mobile rgb-d data, 2022.

\bibitem[Behley et~al.(2019)Behley, Garbade, Milioto, Quenzel, Behnke, Stachniss, and Gall]{Behley2019SemanticKITTIAD}
Jens Behley, Martin Garbade, Andres Milioto, Jan Quenzel, Sven Behnke, C. Stachniss, and Juergen Gall.
\newblock Semantickitti: A dataset for semantic scene understanding of lidar sequences.
\newblock \emph{2019 IEEE/CVF International Conference on Computer Vision (ICCV)}, pages 9296--9306, 2019.

\bibitem[Burri et~al.(2016)Burri, Nikolic, Gohl, Schneider, Rehder, Omari, Achtelik, and Siegwart]{EuROC}
Michael Burri, Janosch Nikolic, Pascal Gohl, Thomas Schneider, Joern Rehder, Sammy Omari, Markus Achtelik, and Roland Siegwart.
\newblock The euroc micro aerial vehicle datasets.
\newblock \emph{The International Journal of Robotics Research}, 35, 2016.

\bibitem[Caesar et~al.(2019)Caesar, Bankiti, Lang, Vora, Liong, Xu, Krishnan, Pan, Baldan, and Beijbom]{Caesar2019nuScenesAM}
Holger Caesar, Varun Bankiti, Alex~H. Lang, Sourabh Vora, Venice~Erin Liong, Qiang Xu, Anush Krishnan, Yuxin Pan, Giancarlo Baldan, and Oscar Beijbom.
\newblock nuscenes: A multimodal dataset for autonomous driving.
\newblock \emph{2020 IEEE/CVF Conference on Computer Vision and Pattern Recognition (CVPR)}, pages 11618--11628, 2019.

\bibitem[Calders et~al.(2022)Calders, Verbeeck, Burt, Origo, Nightingale, Malhi, Wilkes, Raumonen, Bunce, Disney, and M.]{TLS}
Calders, K. Verbeeck, H. Burt, A. Origo, N. Nightingale, J. Malhi, Y. Wilkes, P. Raumonen, P. Bunce, R. Disney, and M.
\newblock Terrestrial laser scanning data wytham woods: leaf-off raw data 2015/16, 2022.

\bibitem[Carlevaris-Bianco~N(2016)]{NCLT}
Eustice~RM Carlevaris-Bianco~N, Ushani~AK.
\newblock University of michigan north campus long-term vision and lidar dataset.
\newblock \emph{The International Journal of Robotics Research}, pages 1023--1035, 2016.

\bibitem[Chang et~al.(2017{\natexlab{a}})Chang, Dai, Funkhouser, Halber, Nießner, Savva, Song, Zeng, and Zhang]{chang2017matterport3d}
Angel Chang, Angela Dai, Thomas Funkhouser, Maciej Halber, Matthias Nießner, Manolis Savva, Shuran Song, Andy Zeng, and Yinda Zhang.
\newblock Matterport3d: Learning from rgb-d data in indoor environments, 2017{\natexlab{a}}.

\bibitem[Chang et~al.(2017{\natexlab{b}})Chang, Dai, Funkhouser, Halber, Nie{\ss}ner, Savva, Song, Zeng, and Zhang]{Chang2017Matterport3DLF}
Angel~X. Chang, Angela Dai, Thomas~A. Funkhouser, Maciej Halber, Matthias Nie{\ss}ner, Manolis Savva, Shuran Song, Andy Zeng, and Yinda Zhang.
\newblock Matterport3d: Learning from rgb-d data in indoor environments.
\newblock \emph{2017 International Conference on 3D Vision (3DV)}, pages 667--676, 2017{\natexlab{b}}.

\bibitem[Chang et~al.(2019)Chang, Lambert, Sangkloy, Singh, Bąk, Hartnett, Wang, Carr, Lucey, Ramanan, and Hays]{Argoverse}
Ming-Fang Chang, John Lambert, Patsorn Sangkloy, Jagjeet Singh, Sławomir Bąk, Andrew Hartnett, De Wang, Peter Carr, Simon Lucey, Deva Ramanan, and James Hays.
\newblock Argoverse: 3d tracking and forecasting with rich maps.
\newblock \emph{2019 IEEE/CVF Conference on Computer Vision and Pattern Recognition (CVPR)}, pages 8740--8749, 2019.

\bibitem[Dai et~al.(2017{\natexlab{a}})Dai, Chang, Savva, Halber, Funkhouser, and Nie{\ss}ner]{dai2017scannet}
Angela Dai, Angel~X Chang, Manolis Savva, Matthew Halber, Thomas Funkhouser, and Matthias Nie{\ss}ner.
\newblock Scannet: Richly-annotated 3d reconstructions of indoor scenes.
\newblock In \emph{Proceedings of the IEEE Conference on Computer Vision and Pattern Recognition (CVPR)}, 2017{\natexlab{a}}.

\bibitem[Dai et~al.(2017{\natexlab{b}})Dai, Chang, Savva, Halber, Funkhouser, and Nie{\ss}ner]{Dai2017ScanNetR3}
Angela Dai, Angel~X. Chang, Manolis Savva, Maciej Halber, Thomas~A. Funkhouser, and Matthias Nie{\ss}ner.
\newblock Scannet: Richly-annotated 3d reconstructions of indoor scenes.
\newblock \emph{2017 IEEE Conference on Computer Vision and Pattern Recognition (CVPR)}, pages 2432--2443, 2017{\natexlab{b}}.

\bibitem[Geiger et~al.(2012)Geiger, Lenz, Stiller, and Urtasun]{geiger2012we}
Andreas Geiger, Philip Lenz, Christoph Stiller, and Raquel Urtasun.
\newblock Are we ready for autonomous driving? the kitti vision benchmark suite.
\newblock In \emph{Proceedings of the IEEE Conference on Computer Vision and Pattern Recognition (CVPR)}, 2012.

\bibitem[Geiger et~al.(2013)Geiger, Lenz, Stiller, and Urtasun]{KITTI}
Andreas Geiger, Philip Lenz, Christoph Stiller, and Raquel Urtasun.
\newblock Vision meets robotics: The kitti dataset.
\newblock \emph{The International Journal of Robotics Research}, 32:\penalty0 1231 -- 1237, 2013.

\bibitem[Gu{\'e}don and Lepetit(2024)]{sun2023sugar}
Antoine Gu{\'e}don and Vincent Lepetit.
\newblock Sugar: Surface-aligned gaussian splatting for efficient 3d mesh reconstruction and high-quality mesh rendering.
\newblock \emph{CVPR}, 2024.

\bibitem[Hackel et~al.(2017)Hackel, Savinov, Ladicky, Wegner, Schindler, and Pollefeys]{hackel2017semantic3d}
T. Hackel, N. Savinov, L. Ladicky, J.D. Wegner, K. Schindler, and M. Pollefeys.
\newblock Semantic3d.net: A new large-scale point cloud classification benchmark.
\newblock In \emph{Proceedings of the IEEE Conference on Computer Vision and Pattern Recognition (CVPR) Workshops}, 2017.

\bibitem[Henrik et~al.(2016)Henrik, R, George, Engin, and B]{DTU}
Aanæs Henrik, Jensen~Rasmus R, Vogiatzis George, Tola Engin, and Dahl~Anders B.
\newblock Large-scale data for multiple-view stereopsis.
\newblock \emph{International Journal of Computer Vision}, 120:\penalty0 153--168, 2016.

\bibitem[Hu et~al.(2022)Hu, Yang, Khalid, Xiao, Trigoni, and Markham]{hu2022sensaturbanlearningsemanticsurbanscale}
Qingyong Hu, Bo Yang, Sheikh Khalid, Wen Xiao, Niki Trigoni, and Andrew Markham.
\newblock Sensaturban: Learning semantics from urban-scale photogrammetric point clouds, 2022.

\bibitem[Iro et~al.(2016)Iro, Ozan, R., Helen, Ioannis, Martin, and Silvio]{s3dis}
Armeni Iro, Sener Ozan, Zamir~Amir R., Jiang Helen, Brilakis Ioannis, Fischer Martin, and Savarese Silvio.
\newblock 3d semantic parsing of large-scale indoor spaces.
\newblock In \emph{Proceedings of the IEEE Conference on Computer Vision and Pattern Recognition (CVPR)}, 2016.

\bibitem[J et~al.(2019)J, Y, Y-S, H, and A]{ComplexUrban}
Jeong J, Cho Y, Shin Y-S, Roh H, and Kim A.
\newblock Complex urban dataset with multi-level sensors from highly diverse urban environments.
\newblock \emph{The International Journal of Robotics Research}, 38(6):\penalty0 642--657, 2019.

\bibitem[Kerbl et~al.(2023)Kerbl, Kopanas, Leimk{\"u}hler, and Drettakis]{kerbl3Dgaussians}
Bernhard Kerbl, Georgios Kopanas, Thomas Leimk{\"u}hler, and George Drettakis.
\newblock 3d gaussian splatting for real-time radiance field rendering.
\newblock \emph{ACM Transactions on Graphics}, 42\penalty0 (4), 2023.

\bibitem[Knapitsch et~al.(2017)Knapitsch, Park, Zhou, and Koltun]{knapitsch2017tanks}
Arno Knapitsch, Jaesik Park, Qian-Yi Zhou, and Vladlen Koltun.
\newblock Tanks and temples: Benchmarking large-scale scene reconstruction.
\newblock \emph{ACM Transactions on Graphics}, 36\penalty0 (4), 2017.

\bibitem[Kulhanek et~al.(2024)Kulhanek, Peng, Kukelova, Pollefeys, and Sattler]{kulhanek2024wildgaussians}
Jonas Kulhanek, Songyou Peng, Zuzana Kukelova, Marc Pollefeys, and Torsten Sattler.
\newblock Wildgaussians: 3d gaussian splatting in the wild.
\newblock \emph{arXiv preprint arXiv:2407.08447}, 2024.

\bibitem[Li et~al.(2023)Li, Jiang, Xu, Xiangli, Wang, Lin, and Dai]{matrixcity}
Yixuan Li, Lihan Jiang, Linning Xu, Yuanbo Xiangli, Zhenzhi Wang, Dahua Lin, and Bo Dai.
\newblock Matrixcity: A large-scale city dataset for city-scale neural rendering and beyond.
\newblock In \emph{Proceedings of the IEEE/CVF International Conference on Computer Vision}, pages 3205--3215, 2023.

\bibitem[Lintong et~al.(2023)Lintong, Michael, Tarimo, David, Marco, Davide, and Maurice]{HiltiOx}
Zhang Lintong, Helmberger Michael, Fu~Lanke~Frank Tarimo, Wisth David, Camurri Marco, Scaramuzza Davide, and Fallon Maurice.
\newblock Hilti-oxford dataset: A millimeter-accurate benchmark for simultaneous localization and mapping.
\newblock \emph{IEEE Robotics and Automation Letters}, 8\penalty0 (1):\penalty0 408–415, 2023.

\bibitem[Liu et~al.(2025)Liu, Luo, Fan, Wang, Peng, and Zhang]{doe2023citygaussian}
Yang Liu, Chuanchen Luo, Lue Fan, Naiyan Wang, Junran Peng, and Zhaoxiang Zhang.
\newblock Citygaussian: Real-time high-quality large-scale scene rendering with gaussians.
\newblock In \emph{European Conference on Computer Vision}, pages 265--282. Springer, 2025.

\bibitem[M. et~al.(2022)M., K., B., N., G., and D.]{Hilti_SLAM_Challenge}
Helmberger M., Morin K., Berner B., Kumar N., Cioffi G., and Scaramuzza D.
\newblock The hilti slam challenge dataset.
\newblock \emph{IEEE Robotics and Automation Letters}, 7\penalty0 (3):\penalty0 7518--7525, 2022.

\bibitem[Matrone et~al.(2020)Matrone, Lingua, Pierdicca, Malinverni, Paolanti, Gilli, Remondino, Mufrsoq, and Landos]{matrone2020arch}
F. Matrone, A. Lingua, R. Pierdicca, E.~S. Malinverni, M. Paolanti, E. Gilli, F. Remondino, A. Mufrsoq, and T. Landos.
\newblock {A BENCHMARK FOR LARGE-SCALE HERITAGE POINT CLOUD SEMANTIC SEGMENTATION}.
\newblock In \emph{International Archives of the Photogrammetry, Remote Sensing and Spatial Information Sciences}, pages 1419--1426, 2020.

\bibitem[PE et~al.(2022)PE, M., JL, P., L., V., O., and M.]{LaMAR}
Sarlin PE, Dusmanu M., Schonberger JL, Speciale P., Gruber L., Larsson V., Miksik O., and Pollefeys M.
\newblock Lamar: Benchmarking localization and mapping for augmented reality.
\newblock In \emph{European Conference on Computer Vision (ECCV)}, pages 686--704, 2022.

\bibitem[Ramezani et~al.(2020)Ramezani, Wang, Camurri, Wisth, Mattamala, and Fallon]{newercollege}
Milad Ramezani, Yiduo Wang, Marco Camurri, David Wisth, Mat{\'i}as Mattamala, and Maurice~F. Fallon.
\newblock The newer college dataset: Handheld lidar, inertial and vision with ground truth.
\newblock \emph{2020 IEEE/RSJ International Conference on Intelligent Robots and Systems (IROS)}, pages 4353--4360, 2020.

\bibitem[Sch{\"o}nberger and Frahm(2016)]{schonberger2016structure}
Jacob~L Sch{\"o}nberger and Jan-Michael Frahm.
\newblock Structure-from-motion revisited.
\newblock In \emph{Proceedings of the IEEE Conference on Computer Vision and Pattern Recognition (CVPR)}, 2016.

\bibitem[Sch{\"o}ps et~al.(2017)Sch{\"o}ps, Sch{\"o}nberger, Galliani, Sattler, Schindler, Pollefeys, and Geiger]{MVS}
Thomas Sch{\"o}ps, Johannes~L. Sch{\"o}nberger, S. Galliani, Torsten Sattler, Konrad Schindler, Marc Pollefeys, and Andreas Geiger.
\newblock A multi-view stereo benchmark with high-resolution images and multi-camera videos.
\newblock \emph{2017 IEEE Conference on Computer Vision and Pattern Recognition (CVPR)}, pages 2538--2547, 2017.

\bibitem[Sch\"ops et~al.(2019)Sch\"ops, Sattler, and Pollefeys]{ETH3D}
Thomas Sch\"ops, Torsten Sattler, and Marc Pollefeys.
\newblock {BAD SLAM}: Bundle adjusted direct {RGB-D SLAM}.
\newblock In \emph{Conference on Computer Vision and Pattern Recognition (CVPR)}, 2019.

\bibitem[Schubert et~al.(2018)Schubert, Goll, Demmel, Usenko, Stuckler, and Cremers]{TUMVI}
David Schubert, Thore Goll, Nikolaus Demmel, Vladyslav Usenko, Jorg Stuckler, and Daniel Cremers.
\newblock The tum vi benchmark for evaluating visual-inertial odometry.
\newblock In \emph{2018 IEEE/RSJ International Conference on Intelligent Robots and Systems (IROS)}. IEEE, 2018.

\bibitem[Smith et~al.(2009)Smith, Baldwin, Churchill, Paul, and Newman]{newcollege}
Mike Smith, Ian Baldwin, Winston Churchill, Rohan Paul, and Paul Newman.
\newblock The new college vision and laser data set.
\newblock \emph{I. J. Robotic Res.}, 28:\penalty0 595--599, 2009.

\bibitem[Straub et~al.(2019)Straub, Whelan, Ma, Chen, Wijmans, Green, Engel, Mur-Artal, Ren, Verma, et~al.]{straub2019replica}
Julian Straub, Thomas Whelan, Lingni Ma, Yufan Chen, Erik Wijmans, Simon Green, Jakob~J Engel, Raul Mur-Artal, Carl Ren, Shobhit Verma, et~al.
\newblock The replica dataset: A digital replica of indoor spaces.
\newblock \emph{arXiv preprint arXiv:1906.05797}, 2019.

\bibitem[Sturm et~al.(2012)Sturm, Burgard, and Cremers]{Sturm2012EvaluatingEA}
J{\"u}rgen Sturm, Wolfram Burgard, and Daniel Cremers.
\newblock Evaluating egomotion and structure-from-motion approaches using the tum rgb-d benchmark.
\newblock 2012.

\bibitem[Sun et~al.(2019)Sun, Kretzschmar, Dotiwalla, Chouard, Patnaik, Tsui, Guo, Zhou, Chai, Caine, Vasudevan, Han, Ngiam, Zhao, Timofeev, Ettinger, Krivokon, Gao, Joshi, Zhang, Shlens, Chen, and Anguelov]{waymo}
Pei Sun, Henrik Kretzschmar, Xerxes Dotiwalla, Aurelien Chouard, Vijaysai Patnaik, Paul Tsui, James Guo, Yin Zhou, Yuning Chai, Benjamin Caine, Vijay Vasudevan, Wei Han, Jiquan Ngiam, Hang Zhao, Aleksei Timofeev, Scott~M. Ettinger, Maxim Krivokon, Amy Gao, Aditya Joshi, Yu Zhang, Jonathon Shlens, Zhifeng Chen, and Dragomir Anguelov.
\newblock Scalability in perception for autonomous driving: Waymo open dataset.
\newblock \emph{2020 IEEE/CVF Conference on Computer Vision and Pattern Recognition (CVPR)}, pages 2443--2451, 2019.

\bibitem[T. et~al.(2023)T., Y., S., S., K., M., and I.]{Nothing_Stands_Still}
Sun T., Hao Y., Huang S., Savarese S., Schindler K., Pollefeys M., and Armeni I.
\newblock Nothing stands still: A spatiotemporal benchmark on 3d point cloud registration under large geometric and temporal change.
\newblock \emph{arXiv preprint arXiv:2311.09346}, 2023.

\bibitem[Tao et~al.(2024)Tao, Munoz-Ban'on, Zhang, Wang, Fu, and Fallon]{OxSpires}
Yifu Tao, Miguel~'Angel Munoz-Ban'on, Lintong Zhang, Jiahao Wang, Lanke Frank~Tarimo Fu, and Maurice~F. Fallon.
\newblock The oxford spires dataset: Benchmarking large-scale lidar-visual localisation, reconstruction and radiance field methods.
\newblock 2024.

\bibitem[Tobias et~al.(2017)]{Semantic3D}
Hackel Tobias et~al.
\newblock Semantic3d dataset for large-scale point cloud classification.
\newblock 2017.

\bibitem[Vijayanarasimhan et~al.(2017)Vijayanarasimhan, Ricco, Schmid, Sukthankar, and Fragkiadaki]{Vijayanarasimhan2017SfMNetLO}
Sudheendra Vijayanarasimhan, Susanna Ricco, Cordelia Schmid, Rahul Sukthankar, and Katerina Fragkiadaki.
\newblock Sfm-net: Learning of structure and motion from video.
\newblock \emph{ArXiv}, abs/1704.07804, 2017.

\bibitem[Wang et~al.(2004)Wang, Bovik, Sheikh, and Simoncelli]{1284395}
Zhou Wang, A.C. Bovik, H.R. Sheikh, and E.P. Simoncelli.
\newblock Image quality assessment: from error visibility to structural similarity.
\newblock \emph{IEEE Transactions on Image Processing}, 13\penalty0 (4):\penalty0 600--612, 2004.

\bibitem[X et~al.(2015)X, A, J, Pat, and Qixing]{ShapeNet}
Chang~Angel X, Funkhouser~Thomas A, Guibas~Leonidas J, Hanrahan Pat, and Huang Qixing.
\newblock Shapenet: An information-rich 3d model repository.
\newblock \emph{arXiv preprint arXiv:1512.03012}, 2015.

\bibitem[Xia et~al.(2018)Xia, Zamir, He, Sax, Malik, and Savarese]{savva2017gibson}
Fei Xia, Amir Zamir, Zhi-Yang He, Alexander Sax, Jitendra Malik, and Silvio Savarese.
\newblock Gibson env: Real-world perception for embodied agents, 2018.

\bibitem[Xiong et~al.(2024{\natexlab{a}})Xiong, Li, and Li]{GauU-Scene}
Butian Xiong, Zhuo Li, and Zhen Li.
\newblock Gauu-scene: A scene reconstruction benchmark on large scale 3d reconstruction dataset using gaussian splatting, 2024{\natexlab{a}}.

\bibitem[Xiong et~al.(2024{\natexlab{b}})Xiong, Li, and Li]{gauu_scene}
Butian Xiong, Zhuo Li, and Zhen Li.
\newblock Gauu-scene: A scene reconstruction benchmark on large scale 3d reconstruction dataset using gaussian splatting, 2024{\natexlab{b}}.

\bibitem[Yadav et~al.(2023)Yadav, Ramrakhya, Ramakrishnan, Gervet, Turner, Gokaslan, Maestre, Chang, Batra, Savva, Clegg, and Chaplot]{habitat}
Karmesh Yadav, Ram Ramrakhya, Santhosh~Kumar Ramakrishnan, Theo Gervet, John Turner, Aaron Gokaslan, Noah Maestre, Angel~Xuan Chang, Dhruv Batra, Manolis Savva, Alexander~William Clegg, and Devendra~Singh Chaplot.
\newblock Habitat-matterport 3d semantics dataset, 2023.

\bibitem[Yeshwanth et~al.(2023)Yeshwanth, Liu, Nie{\ss}ner, and Dai]{scannetplus}
Chandan Yeshwanth, Yueh-Cheng Liu, Matthias Nie{\ss}ner, and Angela Dai.
\newblock Scannet++: A high-fidelity dataset of 3d indoor scenes.
\newblock \emph{2023 IEEE/CVF International Conference on Computer Vision (ICCV)}, pages 12--22, 2023.

\bibitem[Yogamani et~al.(2021)Yogamani, Hughes, Horgan, Sistu, Varley, O'Dea, Uricar, Milz, Simon, Amende, Witt, Rashed, Chennupati, Nayak, Mansoor, Perroton, and Perez]{woodscape}
Senthil Yogamani, Ciaran Hughes, Jonathan Horgan, Ganesh Sistu, Padraig Varley, Derek O'Dea, Michal Uricar, Stefan Milz, Martin Simon, Karl Amende, Christian Witt, Hazem Rashed, Sumanth Chennupati, Sanjaya Nayak, Saquib Mansoor, Xavier Perroton, and Patrick Perez.
\newblock Woodscape: A multi-task, multi-camera fisheye dataset for autonomous driving, 2021.

\bibitem[Yu et~al.(2024)Yu, Sattler, and Geiger]{lee2024dof}
Zehao Yu, Torsten Sattler, and Andreas Geiger.
\newblock Gaussian opacity fields: Efficient adaptive surface reconstruction in unbounded scenes, 2024.

\bibitem[Zhang et~al.(2018)Zhang, Isola, Efros, Shechtman, and Wang]{8578166}
Richard Zhang, Phillip Isola, Alexei~A. Efros, Eli Shechtman, and Oliver Wang.
\newblock The unreasonable effectiveness of deep features as a perceptual metric.
\newblock In \emph{2018 IEEE/CVF Conference on Computer Vision and Pattern Recognition}, pages 586--595, 2018.

\bibitem[Zhirong et~al.(2014)Zhirong, Shuran, Aditya, Xiao, and Jianxiong]{ModelNet40}
Wu Zhirong, Song Shuran, Khosla Aditya, Tang Xiao, and Xiao Jianxiong.
\newblock 3d shapenets: A deep representation for volumetric shapes.
\newblock \emph{arXiv preprint arXiv:1406.5670}, 2014.

\end{thebibliography}


\end{document}